\pdfoutput=1

\documentclass[11pt]{article}

\usepackage[final]{acl}

\usepackage{times}
\usepackage{latexsym}
\usepackage{makecell}
\usepackage{CJKutf8}
\usepackage[T1]{fontenc}

\usepackage[utf8]{inputenc}
\usepackage{booktabs}
\usepackage{enumitem}
\usepackage{multirow}
\usepackage{colortbl}
\usepackage{tikz}
\usepackage{amssymb}
\usepackage{microtype}
\usepackage{soul}
\usepackage{subcaption}
\usepackage{inconsolata}

\usepackage{graphicx}
\usepackage{xspace}
\usepackage{float}
\usepackage{amsmath}
\usepackage{comment}
\usepackage{arydshln}
\usepackage{float,lscape}
\usepackage{afterpage}
\usepackage[colorinlistoftodos,prependcaption,textsize=tiny]{todonotes}

\newcommand{\wz}[1]{\textcolor{black}{#1}}
\newcommand{\se}[1]{\textcolor{black}{#1}}
\newcommand{\zr}[1]{\textcolor{black}{#1}}

\newcolumntype{C}[1]{>{\centering\let\newline\\\arraybackslash\hspace{0pt}}m{#1}}

\title{
How Good Are LLMs for Literary Translation, Really?
\\ 
Literary Translation Evaluation with Humans and LLMs}

\author{Ran Zhang \\
Natural Language Learning Group (NLLG) \\
School of Business Informatics and Mathematics \\
University of Mannheim \\
\texttt{ran.zhang@uni-mannheim.de} 
 \And  ... \And
         Author n \\ Address line \\ ... \\ Address line}
\author{
  \textbf{Ran Zhang\textsuperscript{1,4}}\,
  \textbf{Wei Zhao\textsuperscript{2}}\,
 \textbf{Steffen Eger\textsuperscript{3,4}}
\\
  \textsuperscript{1}University of Mannheim, School of Business Informatics and Mathematics \\
  \textsuperscript{2}University of Aberdeen, Department of Computing Science\\
  \textsuperscript{3}University of Technology Nuremberg (UTN), Dep.~Computer Science \& AI
  \\ \textsuperscript{4} Natural Language Learning and Generation (NLLG) Lab
\\
  \small{
    \textsuperscript{1} \href{mailto:ran.zhang@uni-mannheim.de}{ran.zhang@uni-mannheim.de} \ \ 
    \textsuperscript{2}  \href{mailto:wei.zhao@abdn.ac.uk}{wei.zhao@abdn.ac.uk} \ \ 
    \textsuperscript{3}  \href{mailto:steffen.eger@utn.de}{steffen.eger@utn.de}
  }
}
\newcommand{\liteval}{\textsc{LitEval-Corpus}\xspace}
\newcommand{\deen}{\textit{De-En}\xspace}
\newcommand{\enDe}{\textit{En-De}\xspace}
\newcommand{\enzh}{\textit{En-Zh}\xspace}
\newcommand{\zhen}{\textit{Zh-En}\xspace}
\newcommand{\dezh}{\textit{De-Zh}\xspace}
\newcommand{\gtran}{Google Translate\xspace}
\newcommand{\deepl}{DeepL\xspace}
\newcommand{\gptfo}{GPT-4o\xspace}

\newcommand{\qwen}{Qwen\xspace}
\newcommand{\gemba}{\textsc{GEMBA-MQM}\xspace}

\newcommand{\cometxl}{\textsc{XCOMET-XL}\xspace}
\newcommand{\cometxxl}{\textsc{XCOMET-XXL}\xspace}
\newcommand{\bws}{\textsc{Best-Worst Scaling}\xspace}

\begin{document}
\maketitle
\begin{abstract}
Recent research has focused on literary machine translation (MT) as a new challenge in MT. However, the evaluation of literary MT remains an open problem. We contribute to this ongoing discussion by introducing \liteval, a paragraph-level parallel corpus containing verified human translations and outputs from 9 MT systems, which totals over 2k \zr{translations}
and 13k evaluated sentences across four language pairs, costing 4.5k€. This corpus enables us to (i) examine the \emph{consistency} and \emph{adequacy} of human evaluation schemes with various degrees of complexity, (ii) compare evaluations by students and professionals, assess the effectiveness of (iii) LLM-based metrics and (iv) LLMs themselves.  Our findings indicate that the adequacy of human evaluation is controlled by two factors: the complexity of the evaluation scheme (more complex is less adequate) and the expertise of evaluators (higher expertise yields more adequate evaluations). For instance, MQM (Multidimensional Quality Metrics), a complex scheme and the de facto standard for non-literary human MT evaluation, is largely inadequate for literary translation evaluation: 
\wz{with student evaluators},
nearly 60\% of human translations are misjudged as indistinguishable or inferior to machine translations. In contrast, BWS (\bws), a much simpler scheme, identifies human translations at a rate of \wz{80-100\%}. 
Automatic metrics fare dramatically worse, with rates of at most 20\%. 
Our overall evaluation indicates that published human translations consistently outperform LLM translations, where even the most recent LLMs tend to produce considerably more literal and less diverse translations compared to humans. 

\end{abstract}

\section{Introduction}\label{sec:intro}
\begin{figure}
    \centering
    \includegraphics[width=1\linewidth]{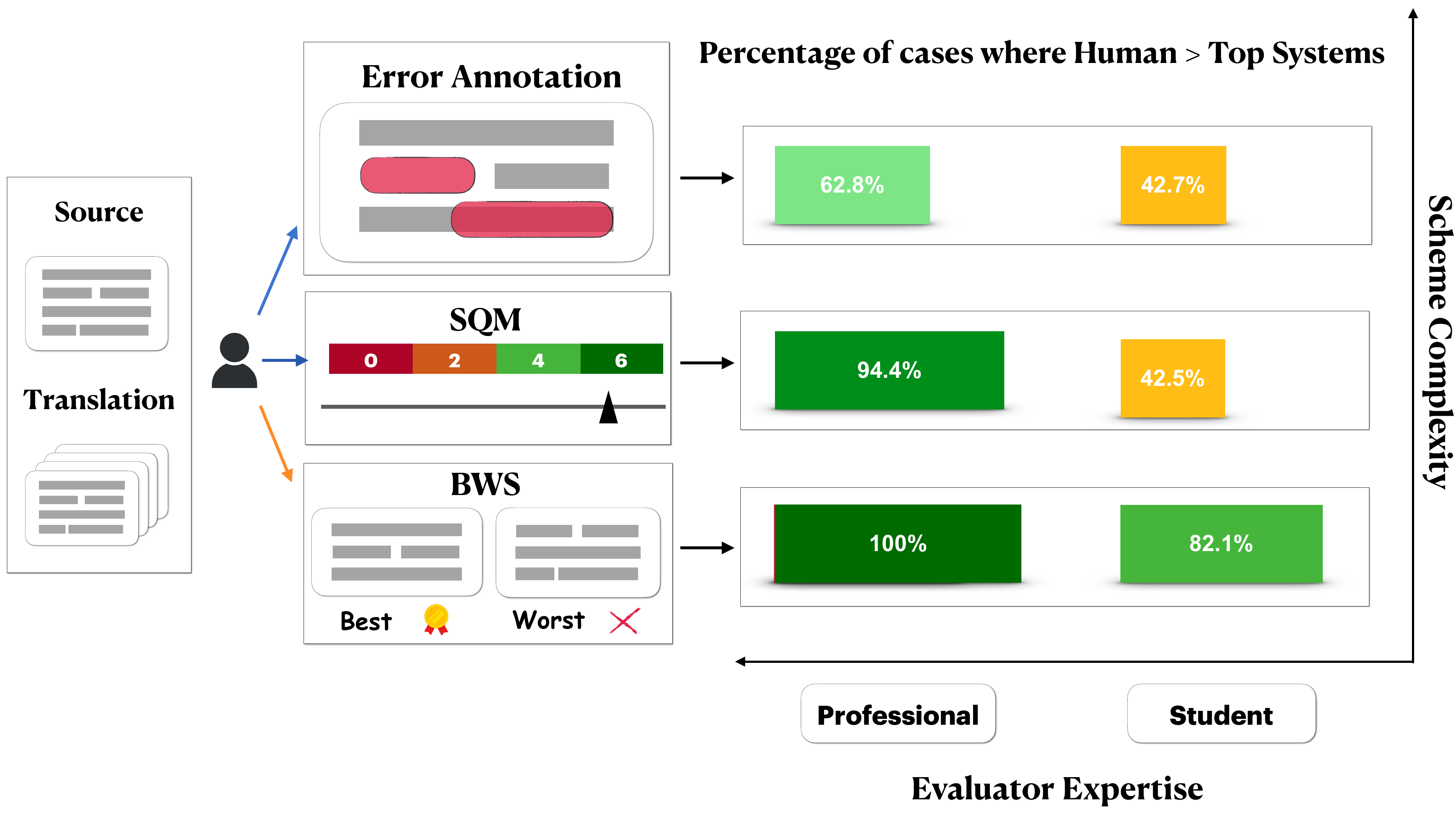}
    \caption{Adequacy comparison per scheme complexity and evaluator expertise. SQM 
    represents Scalar Quality Metrics.
    Professional and student evaluators perform free and MQM 
    error annotation,
    respectively. See Figure \ref{fig:anno_scheme} and Section \ref{sec:preference} for more details.}
    \label{fig:adequacy}
\end{figure}
With the advent of Large Language Models (LLMs), literary translation, once the exclusive domain of human translators \cite{matusov_2019_challenges, hansen2022human, yan2024gpt}, now stands at the threshold of a technological transformation \cite{zhao2023more, wu2024perhaps}. 
LLMs have the potential to increase the volume of translated literary works, particularly for low-resource languages \cite{rutherford2024whence}. 
However, they pose a dual threat: potentially decreasing the salaries of professional translators or even displacing them entirely and diminishing the aesthetic value of translations if LLMs' output quality is persistently overestimated. As a consequence, the field has received much increased attention recently, both in the academic domain \cite{ctt-2024-1, daems-etal-2024-impact, resende-hadley-2024-translators, pang2024salute, wang_2023_findings} and in the translation community \cite{netzwerk, genshagen, goethe, intelligenz}.

Evaluating literary translation poses unique challenges that extend beyond mere linguistic accuracy. It demands a delicate balance of cultural insight, authorial voice, creative interpretation, and aesthetics \cite{voigt_jurafsky_towards, matusov_2019_challenges,macken_2024_LLMmachine}. These nuances surpass the scope of traditional sentence-level MT evaluation methods typically used in non-literary domains, presenting substantial obstacles for existing assessment techniques \cite{van2023riddle}.

We notice the following gaps in existing research regarding literary translation: 
(1) For \textbf{automatic evaluation}, previous metrics like d-BLEU, BLEURT and BLONDE \cite{liu-etal-2020-multilingual-denoising, sellam-etal-2020-bleurt, jiang2022blonde} fall short in discrimination between professional human translations and machine outputs in literary MT \cite{thai2022exploring}. More powerful recent popular metrics such as XCOMET \cite{xcomet_tacl_a_00683}, \gemba \cite{kocmi2023gemba}, and Prometheus \cite{kim2024prometheus} remain untested for literary texts.
(2) The adequacy of \textbf{human evaluation} frameworks such as Multidimensional Quality Metrics (MQM) \cite{freitag_2021_experts, lommel2014using}, Scalar Quality Metric (SQM) \cite{graham-etal-2013-sqm}, and Best-Worst Scaling (BWS) \cite{kiritchenko-mohammad-2017-bestworst} remains largely unexplored: to our knowledge, there is no comprehensive comparison of these schemes for evaluating literary translations. This critical gap is concerning, as current studies apply these approaches arbitrarily, resulting in findings that are not directly comparable and potentially leading to unreliable conclusions. 
(3) Existing \textbf{datasets} are small, cover few LLM models, and lack verification for human translations, including translator qualifications and experience \cite{yan2024gpt}. These issues can lead to substantial misjudgments when comparing LLM outputs to human translations, potentially overestimating the abilities of LLMs to the detriment of human professional translators.

In this paper, we address these gaps with the following key contributions: 
\begin{enumerate}[label=(\roman*)]
\setlength\itemsep{-0.4em}
    \item We introduce \liteval, a paragraph-level parallel corpus with verified high-quality human translations 
    across four language pairs. 
    This corpus also facilitates the comparison of 9 MT systems beyond the GPT series, including commercial models, popular LLMs
    , and previous state-of-the-art (SOTA) MT systems.\footnote{Our dataset and code are publicly accessible on GitHub: \url{https://github.com/zhangr2021/LitMT_eval}.} 
    \item We investigate multiple human evaluation schemes through both students and professional translators, illustrated in Figure \ref{fig:adequacy}, where we find that error annotation including MQM, the de facto measure in standard human MT evaluation, is (at least partially) inadequate for literary translation; SQM's effectiveness for comparing high-quality outputs highly depends on the expertise of evaluators; and BWS proves best for comparing high-quality outputs. 
    \item We examine the effectiveness of popular LLM-based metrics. \gemba consistently outperforms other metrics. However, it primarily focuses on accuracy errors and struggles to differentiate between human translations and literal LLM outputs: it prefers human professional translations over LLM output at a level of 9.6\%, compared to 94.4\% for professional SQM, 81.1\% for 
    student BWS and 42\% for student MQM and SQM. 
    \item We compare system performance, where we find that (1) human translators produce the best translations, as judged by both professionals and students; 
    (2) \gptfo ranks second, followed by \gtran and \deepl. LLM outputs are more literal and less diverse than human translations with recent models decisively better and more closely approaching human literary translation.
\end{enumerate}

\section{Related work}
\paragraph{Human evaluation for literary texts} Literary MT evaluation primarily relies on human judgments of various types. MQM, the widely used guided error annotation scheme in the renowned MT venue WMT \cite{wang_2023_findings, karpinska2023large}, has been shown effective for non-literary texts \cite{freitag_2021_experts} but its suitability for literary translation is uncertain. However, \citet{kocmi-etal-2024-error} indicate that the complex MQM error categorization system requires costly trained experts and may lead to low annotation agreement. BWS is also employed for literary texts, including poetry \cite{thai2022exploring, belouadi_2023_bygpt5}, but has limitations in comparing multiple systems simultaneously \cite{kiritchenko-mohammad-2017-bestworst}. Recently, \citet{wu2024perhaps} claim that LLM-based multi-agent literary translations achieve parity with human translations, based on preference comparison by crowdsourced and LLM evaluator---though the expertise of human translators remains unclear. SQM, a Likert scale rating, is used in non-literary domains. To our knowledge, a comprehensive comparison of these schemes for literary MT is lacking, with different studies employing various methods ad libitum. Our work provides the first systematic comparison of these evaluation methods for literary texts. 

\textbf{Human evaluation} of language models has been the subject of recent debate \citep{thomson-etal-2024-common,belz-etal-2023-missing}. Human evaluation is typically assessed via \emph{consistency}, i.e., whether human annotations are similar. Besides consistency, we focus on \emph{adequacy}: whether the implications of evaluation are plausible (e.g., whether human reference texts are ranked the highest) \cite{toral-etal-2018-attaining}. 

\paragraph{Automatic metrics} for MT have been extensively studied at sentence and document levels in non-literary domains. These include NLI-based MENLI \cite{chen2023menli}, trained metrics like the COMET series \cite{xcomet_tacl_a_00683}, BERT-based scores such as BLEURT \cite{sellam-etal-2020-bleurt} and BERTScore \cite{zhangbertscore}, prompting-based metrics like \gemba \cite{kocmi2023gemba} and MQM-APE \cite{lu2024mqm}, in addition to document-level BLONDE \cite{jiang2022blonde} and DiscoScore \cite{zhao2023discoscore}. While some researchers have applied these metrics to literary texts \cite{wang_2023_findings, hansen2022human}, their effectiveness in literary domain remains understudied. Previous work has examined BLEU, BLEURT, and BLONDE for identifying top-performing systems \cite{van2023riddle, thai2022exploring}, with limited correlation analysis (e.g., brief discussions in appendices) of COMET, COMET-QE, BLEURT, and BERTScore \cite{karpinska2023large}. Our study addresses this gap by evaluating the effectiveness and adequacy of recent LLM-based metrics for literary translation. 

\paragraph{Corpora for literary MT evaluation}
Existing parallel corpora are insufficient for literary translation evaluation. Table \ref{tab:related_corpus} (appendix) summarizes the statistics. 
\textit{BWB} \cite{jiang2022blonde} and \textit{GuoFeng} \cite{xu2022guofeng} focus on recent \zhen web novels with unclear translation origins, potentially post-edited MT outputs \cite{kolb2023bit}. \textit{PAR3} \cite{thai2022exploring} lacks crucial meta-information, such as the origin of human translations, and offers limited evaluation with small-scale A/B tests comparing outputs from only three generation systems. \citet{karpinska2023large} compare two systems (\gtran and GPT-3.5) with only contemporary works---usually easier to translate than classics and potentially involving post-edited MT. Our \liteval addresses these limitations, containing over 2k paragraph-level segments with 13k sentences, comparing nine systems against verified published human translations with human evaluation under multiple schemes. 

\section{
\liteval} 
\label{sec:dataset}
We detail our \liteval in this section, covering data construction and statistics.
\subsection{Dataset construction}
Our data construction process consists of 3 steps: (1) preselect and construct paragraph-level parallel corpus; (2) select MT systems and collect translation hypotheses; 
(3) assess translation quality.

\paragraph{Constructing parallel corpus}
We consider three languages: 
English, Chinese, and German, covering translations between four different language pairs, i.e.,
English-Chinese (\enzh), German-English (\deen)/English-German (\enDe), and German-Chinese (\dezh).\footnote{The authors are native/proficient in these three languages and can thus verify the quality of the 
corpus and the evaluation.} We begin by collecting paragraph-level parallel corpus for each language pair, including both classic and contemporary works. Classics present more challenges due to their complex syntax, language change, and cultural references that may no longer be familiar to modern audiences. Notably, older translations are especially valuable, as they were created before the widespread use of MT and are thus free from concerns of being post-edited MT outputs. Additionally, including contemporary works helps mitigate potential data contamination in LLMs, as these models are more likely to have been pre-trained on widely available texts, including classic literature. 

To ensure the quality of our dataset, we derive both source and target paragraphs from verified publications, including books from Project Gutenberg \cite{stroube2003literary}, reading samples from published works, and purchased materials. For classic works, the target paragraphs include at least two human translations, while contemporary works have at least one. For classic \deen works, we utilize PAR3 samples \cite{thai2022exploring}.\footnote{PAR3 includes 9k \deen paragraphs aligned with human translations but has alignment issues for some paragraphs with unidentified translation origins. To address these, the first author proofread sampled source-translation pairs and verified the origin of each human translation.} For contemporary \deen works and the other three language pairs (\dezh, \enzh, and \enDe), we build the corpus from scratch, where the first author manually aligns source and translation paragraph pairs.

\paragraph{Selecting MT systems and collecting translation hypotheses}

We explore various-sized models, including cost-efficient alternatives to larger models:
\begin{itemize}
\setlength\itemsep{-0.4em}
\item \textbf{Commercial systems} like \gtran and \deepl. \deepl is particularly noteworthy for translators due to its widespread use within the translation community.
\item Previous SOTA \textbf{transformer models} optimized for sentence-level translation, including NLLB-3.3b \cite{costa2022nllb} and M2M\_100-1.3b \cite{m2m2021}.
\item Closed-source \gptfo and four series of open-source \textbf{LLMs} of smaller sizes:\footnote{For detailed parameter size, see Table \ref{tab:systranking} (appendix).} (1) Llama 3 \cite{llama3modelcard}, one of the strongest open-source models; (2) Qwen 2 \cite{yang2024qwen2}, specialized in Chinese; (3) Gemma 1.1 \cite{team2024gemma}, developed with the same technology as the closed-source Gemini model \cite{team2023gemini}; and (4) TowerInstruct \cite{tower_llm_2024}, trained for translation-related tasks. We use identical prompts across all LLMs when requesting literary translations: \textit{``Please translate the following literary texts from [source language] to [target language]. The texts are as follows: [texts]''}.\footnote{We explored multiple prompts but observed no clear superiority of other prompts over the basic one. For more details, see Section \ref{sec:prompt} (appendix).} 
\end{itemize}

\subsection{Assessing translation quality}
Our study examines three human evaluation schemes, i.e., guided/free error annotation, SQM, and BWS, using both student evaluators and professional translators. Figure \ref{fig:anno_scheme} (appendix) summarizes the evaluation schemes. For \textbf{guided error annotation (MQM)}, evaluators read the source/candidate translation pair, identify error spans, and categorize them based on the annotation guideline with six major categories: (1) Terminology; (2) Accuracy;
(3) Fluency; 
(4) Style;
(5) Locale-convention; and (6) Non-translation.\footnote{See Section \ref{guideline} (appendix) for annotation details.} 
Each error is then labeled as major or minor. The MQM score is thus computed with the following formula: \[-\frac{
(C_{\text{Non-translation}} \times 25 + C_{\text{major}} \times 5 + C_{\text{minor}} \times 1)}{\text{number of sentences per paragraph}}\] $C$ represents the count of errors. In contrast, \textbf{free error annotation} does not follow any guidelines. The evaluators highlight spans indicating erroneous/good translations and explain their highlights. We 
can thus compute span-level agreement with student error annotation.\footnote{Our feedback from translators revealed that MQM error annotation is unfamiliar to the translation community. Free error annotation better reflects the editing process and enables us to assess MQM protocol by comparing error span matches.}
For \textbf{SQM}, evaluators provide a scale rating between 0-6 for the overall quality of the translation, especially the translation's stylistic and artistic qualities based on the source. Evaluators may scroll up and down to check all other translations for scoring comparison. \textbf{BWS} is a direct comparison method that identifies the best and worst translations. 

\paragraph{Evaluator statistics} 
We involve four student evaluators, one for each language pair, and four professional translators (one for \enzh, two for \enDe, and one for \deen). 
All evaluators must be native speakers of the target language. Three students are master students with bachelor degrees in linguistics (\dezh), computational linguistics (\deen and \enDe) and one senior bachelor in translation studies (\enzh). The student evaluators for \dezh, \deen, and \enzh are female and the \enDe evaluator is male. For professional evaluators, we recruit two literary translators with verified publication records for ten hours each. The female translator for \enzh has published a translation of a classic work with a public rating of 9.3/10. The female \enDe translator is highly experienced, having published over 30 translated works. The other two translators volunteer to evaluate for six hours each. Both are female, recruited by universities, and highly experienced in academic translation.

We divide our human evaluation into three groups due to budget constraints:
\begin{itemize}
\setlength\itemsep{-0.4em}
\item Students conduct MQM and SQM evaluation for \textit{all segments} using Label Studio platform \cite{Labelstudio}.
\item The same students perform BWS on \textit{samples}, comparing translations from human translators, \gptfo, \deepl, and \gtran (\qwen for pairs involving Chinese) blindly.\footnote{Evaluating all systems in one entry with 10-12 segments may decrease the reliability of BWS \cite{kiritchenko-mohammad-2017-bestworst}. We include 4-5 translation segments from the top-performing systems to achieve the best outcome.}
\item Professional translators perform free error annotation and SQM on \textit{samples} from human translator\se{s}, \gptfo, \gtran, and \deepl (\qwen for pairs involving Chinese).\footnote{We focus on segments from top systems per feedback from professionals. Other translations often contain major issues, making them less valuable for professional assessment (e.g., highlighting several problematic sentences).}
\end{itemize}

\subsection{\liteval statistics}
Table \ref{tab:statistics} summarizes the statistics of \liteval. The dataset contains 188 \zr{source} paragraphs, encompassing 2,188 translation segments at paragraph level including over 13k sentences. On average, source paragraphs contain 5.6 sentences, while target paragraphs have 6.1 sentences. To our knowledge, \liteval is the most extensive dataset for literary translation evaluation, offering comprehensive information on both source and target translations. It uniquely features paragraph-level translations across 10 generation systems.\footnote{See Table \ref{tab:meta} (appendix) for information on source and translation details.}
\begin{table}[H]
  \centering
\scriptsize
\begin{tabular}{lcccccc}\toprule
\multirow{2}{*}{\textbf{pair}} &\multirow{2}{*}{\textbf{\#paragraph}} &\multirow{2}{*}{\textbf{\#segments}} &\multirow{2}{*}{\textbf{\#sentences}} &\multicolumn{2}{c}{\textbf{\#sent/para }} \\\cmidrule{5-6}
& & & &\textbf{source} &\textbf{target}
\\\midrule 
\textbf{\deen} &46 &562 &4310 &7.1 &7.7 \\
\textbf{\enDe} &46 &554 &2790 &4.4 &5.0 \\
\textbf{\dezh} &47 &520 &3039 &5.3 &5.8 \\
\textbf{\enzh} &49 &552 &3162 &5.5 &5.7 \\
\hdashline
\textbf{Total} &188 &2188 &13301 &5.6 &6.1 \\
\arrayrulecolor{black!100}\bottomrule
\end{tabular}
  \caption{\label{tab:statistics}
   Characteristics of \liteval. \#sentences and \#sent/para represent the total number of annotated sentences in translations and the number of sentences per paragraph, respectively. \#paragraph and \#segments represent the number of source paragraphs and translation segments from all systems, respectively.
  }
\end{table}

\section{Human evaluation} 
We use \liteval to examine the consistency and adequacy of different schemes.
\subsection{Consistency}
\paragraph{Student vs.\ student} 
\begin{table}
\centering
\scriptsize
\begin{tabular}{lccccc:c}\toprule
\multirow{2}{*}{Pair} &\multirow{2}{*}{MQM} &\multirow{2}{*}{SQM} &\multicolumn{2}{c}{Error span match} &\multirow{2}{*}{\#segments} &\multirow{2}{*}{BWS}
\\\cmidrule{4-5}
& & &span &label & & \\\midrule
\textbf{\deen} &0.464 &0.455 &0.308 &0.243 &45 &0.578 \\
\textbf{\enDe} &0.434 &0.350 &0.348 &0.283 &53 &0.564 \\
\textbf{\dezh} &- &- &- &- &- &- \\
\textbf{\enzh} &0.582 &0.656 &0.452 &0.220 &33 &0.581 \\
\hdashline
\textbf{Mean} &0.493 &0.487 &0.369 &0.249 &- & 0.574 \\
\arrayrulecolor{black!100}\bottomrule
\\
\multicolumn{7}{c}{(a)}
\\
\\
\end{tabular}
\begin{tabular}{lcccccc}
\toprule
\multirow{2}{*}{Pair} &\multicolumn{2}{c}{SQM} & &\multicolumn{2}{c}{\makecell{Error span match\\(span)}} &\multirow{2}{*}{\#segments} \\\cmidrule{2-3}\cmidrule{5-6}
&S-S &S-P & &S-S &S-P & \\
\midrule
\textbf{\deen} &0.528 &0.363 & &0.282 &0.252 &20 \\
\textbf{\enDe} &0.151 &0.196 & &0.442 &0.318  &26 \\
\textbf{\dezh} &-&-&&-&-&- \\
\textbf{\enzh} &0.359 &0.517 & &0.374 &0.297 &22 \\
\hdashline
\textbf{Mean} &0.346 &0.359 & &0.366 &0.289  & -\\
\arrayrulecolor{black!100}\bottomrule\\
\multicolumn{7}{c}{(b)}
\end{tabular}
\caption{Annotation agreement (a) between pairs of two student evaluators (\textbf{S-S}); (b) between student and professional translators (\textbf{S-P}). The agreement for MQM score and SQM score is measured by Kendall’s Tau; the agreement for BWS and the span-level comparison is measured by Cohen’s kappa, where span match reflects the agreement on error spans regardless of labels, and label match on both spans and labels.}\label{tab:agree_student}
\end{table}

\input{}
Table \ref{tab:agree_student} (a) shows the annotation agreement between pairs of student evaluators across three language pairs: \deen, \enDe, and \enzh.\footnote{\deen and \enDe evaluators annotate shared segments for both language pairs. A co-author, a native Chinese speaker with a double major in linguistics and fluent in English, evaluates \enzh samples. For \dezh, we lack additional qualified evaluators.} We assess MQM and SQM using Kendall's Tau \cite{kendall}. For BWS and annotated error spans, we employ Cohen's kappa \cite{mchugh2012interrater}. A span match indicates agreement on error spans regardless of labels, and a label match considers agreement on both spans and labels.\footnote{BWS compares 4-5 systems with samples of 15, 15, 23, and 22 paragraphs for \deen, \enDe, \enzh, and \dezh, respectively. These correspond to 67, 65, 118, and 111 translation segments in total.}

The results show \textit{moderate to high agreement for both MQM and SQM}. MQM agreement ranges from 0.434 for \enDe to 0.582 for \enzh, while SQM scores the highest agreement for \enzh at 0.656 and the lowest for \enDe at 0.350. On average, MQM yields an agreement of 0.493, similar to SQM of 0.487.

In terms of error span agreement, \enzh achieves the highest span match at 0.452, while \enDe shows the highest label match at 0.283. Although error span agreement is generally \textit{lower than MQM and SQM, the results remain competitive} with previous sentence-level multi-domain findings \citep{blain2023findings, freitag_2021_experts, leiter2023eval4nlp}. However, a direct comparison of agreement on paragraph-level annotations in literary domain has yet to be explored, underscoring the value of this study as a baseline for future research in literary translation evaluation.

Additionally, \textit{BWS results indicate strong performance}, with \enzh achieving the highest agreement at 0.581, followed by \deen at 0.578 and \enDe at 0.564 measured by Cohen's kappa. The overall mean BWS agreement is 0.574, showing a consistent agreement in selecting the best and worst translations. 

\paragraph{Professional translators vs.\  student evaluators}
Table \ref{tab:agree_student} (b) compares the annotation agreement between student and professional evaluators (S-P). For comparability, we also report the agreement between student evaluators (S-S) computed with the same segments. For SQM, S-P agreement is on average slightly higher than S-S agreement. For \enzh, the agreement is 0.359 (S-S) versus 0.517 (S-P). One exception is \deen pair, where S-S agreement is 0.528, while S-P agreement drops to 0.363. The agreement for \enDe is low for both groups. For error span match, the agreements between S-S and S-P both decline. Overall, S-S shows better agreement than S-P. This is understandable as student evaluators are trained with an annotation guideline, while professional translators perform free annotation without guidelines. The results indicate that both student and professional evaluators focus on joint error spans to some degree.\footnote{See Section \ref{sec:example} for detailed examples.} While there is \textit{consistency between the two groups}, particularly in certain language pairs, \textit{discrepancies exist}. We discuss these in detail in Section \ref{sec:discre}.

\paragraph{Consistency in system ranking}
The evaluation results in Table \ref{tab:systranking} (appendix) show strong consistency in system rankings between student and professional evaluators for both MQM and SQM. Human translation consistently ranks first in both scores when averaged across all segments in the four languages studied. This top ranking holds for both student and professional evaluators, indicating human translation's superiority. \gptfo ranks second, followed by \gtran and \deepl or \qwen. Smaller models such as M2M and NLLB consistently rank low. Additionally, according to student evaluators, there is a clear performance gap between the top 4 systems and the rest, i.e., 5.5 points drop from the fourth to fifth model in MQM and 1.1 points by SQM. Notably, professional translators rate human translations substantially higher than the second-place \gptfo with a difference of 1.8 points in SQM, while this gap is comparatively smaller with 0.4 in student SQM.

\subsection{Adequacy}\label{sec:discre}
\begin{table*}
  \centering
\scriptsize
\begin{tabular}{ccccccccccccccc}\toprule
\multirow{3}{*}{\textbf{pair}} &\multicolumn{8}{c}{\textbf{\makecell{Human > Top Systems \\ (\gptfo, \deepl, \gtran, and \qwen)}}} & &\multicolumn{5}{c}{\textbf{\makecell{Human > Other Systems \\ (excluding Top systems)}}} \\
\cmidrule{2-9}\cmidrule{11-15}
&\multicolumn{3}{c}{\textbf{Student}} &\textbf{} &\textbf{Professional} &\textbf{} &\multicolumn{2}{c}{\textbf{\gemba }} &\textbf{} &\multicolumn{2}{c}{\textbf{Student}} &\textbf{} &\multicolumn{2}{c}{\textbf{\gemba }} \\\cmidrule{2-4}\cmidrule{6-6}\cmidrule{8-9}\cmidrule{11-12}\cmidrule{14-15}
&\textbf{MQM} &\textbf{SQM} &\textbf{BWS} &\textbf{} &\textbf{SQM} &\textbf{} &\textbf{Original} &\textbf{Literary} &\textbf{} &\textbf{MQM} &\textbf{SQM} &\textbf{} &\textbf{Original} &\textbf{Literary} \\\midrule
\textbf{\deen} &60.0\% &60.0\% &66.7\% & &100.0\% & &6.7\% &0.0\% & &86.7\% &80.0\% & &26.7\% &33.3\% \\
\textbf{\enDe} &35.0\% &60.0\% &86.7\% & &83.3\% & &0.0\% &10.0\% & &95.0\% &100.0\% & &70.0\% &60.0\% \\
\textbf{\dezh} &30.0\% &25.0\% &95.0\% & &- & &15.0\% &20.0\% & &95.0\% &90.0\% & &45.0\% &50.0\% \\
\textbf{\enzh} &45.8\% &25.0\% &80.0\% & &100.0\% & &0.0\% &8.3\% & &87.5\% &79.2\% & &41.7\% &50.0\% \\
\hdashline
\textbf{Mean} &42.7\% &42.5\% &82.1\% & &94.4\% & &5.4\% &9.6\% & &91.0\% &87.3\% & &45.8\% &48.3\% \\
\arrayrulecolor{black!100}\bottomrule
\end{tabular}
  \caption{\label{tab:humanvsllm}
   Percentage of segments where human translations are preferred over machine translations per human evaluation schemes and \gemba versions.}
\end{table*}
We assess the adequacy of translation evaluation schemes by evaluating their ability to differentiate high-quality human translations from machine outputs. We believe that experienced human translators outperform MT systems, based on recent studies for other domain texts \cite{yan2024gpt} and the inherent advantages of professional translators: access to full-text context, professional education, and the ability to research cultural backgrounds. These factors enable translators to develop a deeper understanding of the source material, intuitively resulting in more accurate and creative translations.

Table \ref{tab:humanvsllm} shows the percentage of segments where human translations are preferred over other systems. The table presents two scenarios: (1) human translations compared to top-performing models like \gptfo, \deepl, \gtran, and \qwen, and (2) human translations compared to other MT systems, excluding top-performing models. 

While student \textbf{MQM} demonstrates good performance in scenario (2), the bad performance in scenario (1) indicates that \textit{MQM alone may not be adequate for evaluating translation quality effectively, especially for top systems}. The first reason is that MQM requires complex categorization and expects high expertise. This is evidenced by the inferior consistency score for label match in Table \ref{tab:agree_student} (a) from student evaluators. The second reason is that MQM may be inherently unsuitable for literary translation evaluation. MQM struggles with evaluating top models. For example, Figure \ref{fig:errordistibution} (appendix) illustrates that the error distribution of top models closely resembles that of human translation. 16.6\% of \gptfo outputs are ``error-free'' but still fall short in wording or overall style compared to human translations. Professional translators also express concerns about the ``error-counting'' schemes, particularly when assessing high-quality literary translations, as they sometimes make intentional choices, such as omitting certain elements or slightly altering meanings, to enhance the expression in the target language.\footnote{See examples in Table \ref{tab:mis_label} (appendix).}

On average, \textbf{SQM} with students performs similarly to MQM but professionals show a substantially higher preference for human translation (100\% in \deen and \enzh). Professional SQM outperforms student SQM by a large margin of nearly 40 points, because students---lacking sufficient competence or experience---may struggle to capture nuances through SQM rating as effectively as professionals do. This indicates that SQM's effectiveness heavily depends on the evaluator's experience and expertise. However, combining student MQM and SQM can empirically improve the adequacy by a maximum margin of 20 points for \deen and nearly 5 points for \enDe, \enzh and \dezh compared to SQM or MQM alone (see Figure \ref{fig:appdx_adequacy} in Section \ref{sec:appdix_adequacy} for more details). 

\textbf{BWS}, on the other hand, consistently shows high ratios among top systems, with students preferring human translations at 86.7\% for \enDe and 95.0\% for \dezh. On average, BWS ranks second with 82.1\% compared to $\sim$94\% for professional SQM and $\sim$42\% for student SQM and MQM. This suggests that BWS is more suitable for evaluating top systems, although it cannot provide the detailed error insights that MQM offers. \\

\noindent\textbf{Summary:} Student and professional evaluators consistently evaluate literary translations using guided/free error annotation and SQM at a moderate level with BWS showing slightly better agreement. Student MQM is not adequate to discriminate human translations from top LLM translations. Using it may yield false conclusions. The effectiveness of SQM heavily depends on the expertise of the evaluator. BWS excels at distinguishing human translations from top machine translations even among less-experienced evaluators, though this scheme offers less detailed feedback for further improvements such as post-editing.
\section{Automatic metrics} 
We use \liteval to assess recent automatic metrics. While human MQM and SQM may not fully capture the nuances of literary translation quality, as discussed in the previous section, it is still valuable to examine how consistently current metrics align with human MQM/SQM evaluations. This is particularly relevant given that the recent top-performing metrics are built on similar principles. We focus on reference-free metrics due to the scarcity of reference translations for the majority of literary works \cite{rutherford2024whence} and measure the performance by segment-level correlation between automatic metrics and human evaluation.

We examine four SOTA automatic metrics that rely on LLMs: Prometheus 2 (7 billion parameters) \cite{kim2024prometheus}, an open-source model trained to evaluate other language models, which is queried using a tailored prompt (shown in Table \ref{tab:Prometheus} in the appendix) for SQM evaluation of literary translations; \cometxl/\cometxxl (3.5 and 10 billion parameters) \cite{xcomet_tacl_a_00683}, one of the strongest open-source MT metrics for standard MT, which is fine-tuned to assess quality\se{,} generating scores and error spans with severity labels;\footnote{We apply \cometxl and \cometxxl directly to paragraph-level evaluation, based on \citet{deutsch-etal-2023-training}'s finding that applying sentence-level metrics to paragraphs produces results similar to averaging sentence-level scores. }
\gemba \cite{kocmi2023gemba}\se{,} a top-performing prompting-based metric, which is designed to detect translation quality errors based on the MQM framework using LLMs.\footnote{We utilize \gptfo mini for \gemba in this study due to cost reasons. See Section \ref{sec:autometric} and \ref{sec:gemba_detail} (appendix) for more details.} We modify \gemba with the same error categories we use for student evaluators. 
We implement \gemba (Original) using the original prompt template and \gemba (Literary) adapted with domain knowledge and literary translation examples shown in Table \ref{tab:gembatemplate}. 

\begin{figure}
    \centering    \includegraphics[width=0.95\linewidth]{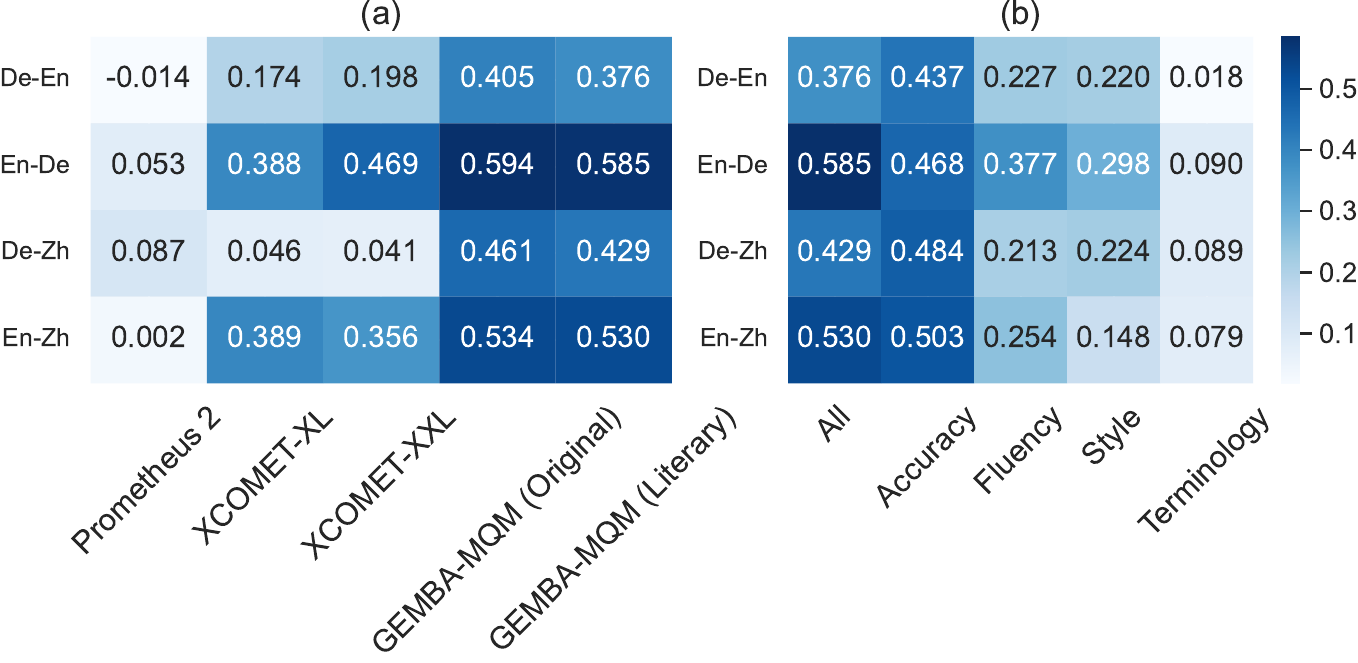}
    \caption{Segment level correlation measured by Kendall’s $\tau$ between (a) human MQM and automatic metrics and (b) human MQM and \gemba (Literary) per error categories. See Figure \ref{fig:sqm_corr} (appendix) for SQM results.}
    \label{fig:corraspect}
\end{figure}
\paragraph{The best metric (\gemba) correlates moderately with human MQM but it cannot distinguish human translation from LLM outputs.} 
Figure \ref{fig:corraspect} (a) illustrates the segment-level correlation between human MQM and automatic metrics. \gemba, particularly its Original version, consistently outperforms other metrics. \gemba (Literary) performs similarly to \gemba (Original), though slightly worse across the board. \cometxl and \cometxxl models rank second overall, demonstrating poor to moderate correlation with human evaluation. Their performance notably declines for more distant language pairs, such as \dezh, possibly due to differences or limitations in their training data for this language direction. Moreover, we notice that \cometxl and \cometxxl are negatively correlated with the length of the target translation at -0.420 and -0.393, respectively. In contrast, the correlations between the length of the target and other metrics are all below $0.05$ in absolute value. Additionally, Prometheus performs the worst, assigning a score of 4 (equivalent to ``good'' translation) to 54.9\% of segments. Its correlation to human SQM ranks lowest among all automatic metrics, as shown in Figure \ref{fig:sqm_corr} (appendix). 

\paragraph{Even the best automatic metric is not adequate for literary MT evaluation.}
\gemba, despite moderate correlation with human SQM and MQM, faces adequacy issues in literary translation evaluation. Table \ref{tab:humanvsllm} shows the percentage of segments where \gemba prefer human translations over MT. The results show \gemba severely struggles to distinguish human translations from top LLM systems. For scenario (1), human translator vs.\ top systems, \gemba (Literary)'s highest percentage (9.6\%) is still 32.9 points lower than the worst human performance. This may partly stem from LLMs' tendency to act as narcissistic evaluators, favoring their own outputs---especially in reference-free scenarios \cite{liu-etal-2024-llms-narcissistic}. 
In scenario (2), human translator vs.\ systems excluding top ones, despite higher percentages, \gemba still lags behind human evaluators by at least 39 points. \gemba (Literary) slightly outperforms the original version by an average of 4.2 points and 2.5 points for scenarios (1) and (2), indicating that literary-specific knowledge helps but remains insufficient for discriminating human and LLM outputs. As shown in Table \ref{tab:humanvsllm_metric} (appendix), other metrics also exhibit substantial gaps compared to human performance. \cometxl scores 26.9 points lower and \cometxxl 21.2 points lower than the worst human performance in scenario (1). In scenario (2), the gaps are even larger with 39.6 points for \cometxl and 30.2 for \cometxxl. While \cometxl and \cometxxl yield better results than \gemba on adequacy, with average performances of 15.6\% and 21.3\% in scenario (1) respectively, they correlate worse with human MQM and SQM than \gemba. This is particularly evident for \deen and \dezh, as shown in Figure \ref{fig:corraspect} (a) and Figure \ref{fig:sqm_corr} (appendix). 

Additionally, our analysis reveals that \gemba's performance is primarily driven by Accuracy aspects, with less reliability in evaluating Fluency, Style, and Terminology (see Figure \ref{fig:corraspect} (b)).\footnote{We aggregate human MQM and \gemba per error category and then compute the correlation separately per each category.} Accuracy shows the strongest correlations across all language pairs. Accuracy for \deen and \dezh even surpasses the full correlation. Meanwhile, Fluency and Style demonstrate poor to fair correlations. Notably, Terminology consistently shows near-zero correlations, which could be problematic for distant language pairs with more culture-specific terms, e.g., \enzh with Terminology as the second most frequent error category. These findings suggest that future metric development should focus on improving the assessment of Fluency, Style, and Terminology. 

\begin{figure}
    \centering
    \includegraphics[width=0.75\linewidth]{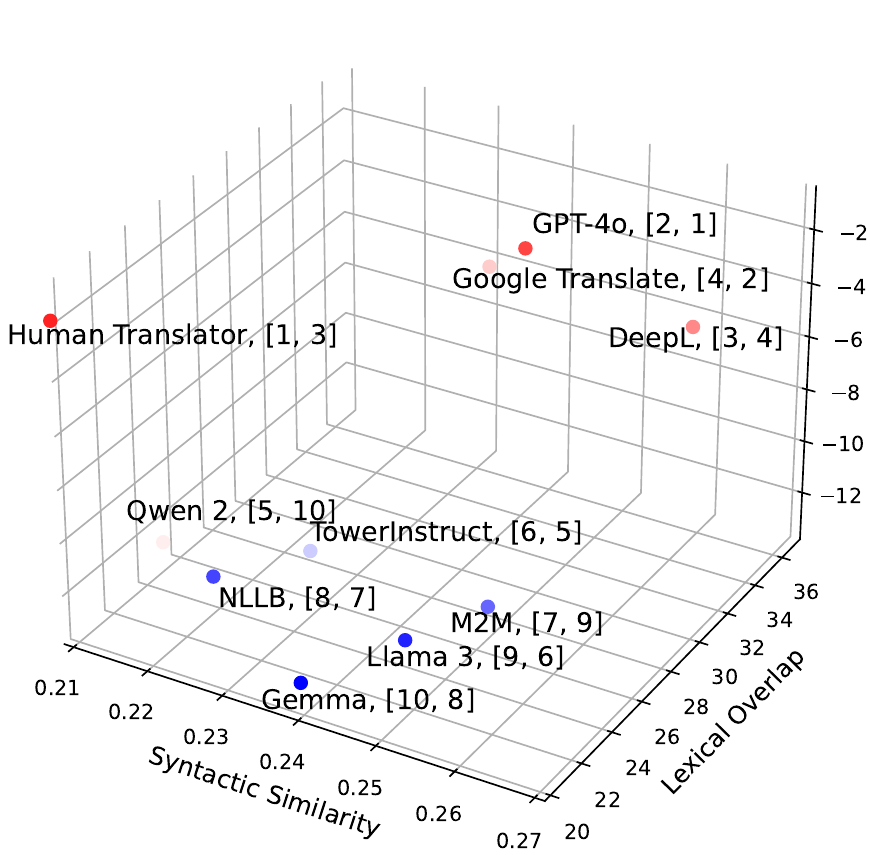}
    \caption{Distribution of human MQM (z-axis) against syntactic similarity between source-translation pairs and lexical overlap of the current system to all other systems. The system rankings are shown as [rank per human MQM, rank per \gemba (Literary)]. See Figure \ref{fig:overlap} for a detailed heatmap.}
    \label{fig:syntacticsym}
\end{figure}

\paragraph{Human MQM vs.\ \gemba: LLMs demonstrate a tendency towards more literal and less diverse translations.}
We further compare human MQM with \gemba in system rankings considering characteristics of the translation systems. To assess the literalness of translation, we analyze the syntactic similarity between source-translation pairs. We utilize the recent \textit{document-level} tree kernel-based syntactic similarity metric FastKASSIM \cite{chen_2023_fastkassim}. While high syntactic similarity preserves the source's form, it frequently sacrifices naturalness in the target language and hinders creativity in translations \cite{guerberof2022creativity, al2021translation}. Diversity is another concern in literary domain \cite{chen2024evaluating, zhang2024llm}. To measure lexical diversity of one system compared to all other systems, we calculate pairwise lexical overlap between pairs of translations from different systems and average them per system.\footnote{See Section \ref{sec:formula_diversity} (appendix) for more details.} A lower value suggests a more diverse vocabulary selection. Figure \ref{fig:syntacticsym} shows results based on \textbf{all publications}, illustrating human MQM scores for 10 translation systems averaged over four language pairs (z-axis) against syntactic similarity (x-axis) and lexical overlap (y-axis) (see Figure \ref{fig:syntacticsym_recent} in the appendix for results on \textbf{contemporary works}). The rank of systems is marked as [rank per human MQM, rank per \gemba]. We show detailed examples in Table \ref{tab:syntax_example1} and Table \ref{tab:syntax_example2} (appendix). 

Three major clusters emerge:
(1) Human Translator stands alone, characterized by a high MQM score with the lowest syntactic similarity (0.21) and lexical overlap (18.9).
While human translation ranks first per human MQM by a 0.4 margin over \gptfo, \gemba (Literary) prefers \gptfo with a larger margin of 1.4 over human translators. Surprisingly, \gemba (Literary) even rates \gtran higher than human translations.
(2) \gptfo, \deepl, and \gtran all exhibit high MQM scores and high lexical overlap. The three models share high lexical overlap, contributing to their remarkably high average scores. \deepl has the highest syntactic similarity of 0.27.
(3) The third cluster demonstrates a performance gap to other clusters, with \qwen 2 and TowerInstruct performing slightly better. Qwen 2 paraphrases more for \dezh and \enzh pairs and sometimes includes Chinese characters in \deen and \enDe outputs, which results in low syntactic similarity and lexical overlap but still higher than human. \\ 

\noindent\textbf{Summary:} \gemba outperforms other current automatic metrics but struggles with non-Accuracy aspects. All metrics lack adequacy in distinguishing human and LLM translations. LLMs tend to produce more literal and less diverse translations showing higher syntactic similarity to the source and greater lexical overlap with other systems than human translations. Further, LLM as evaluator prefers more literal translations and is biased towards their outputs. 

\section{Conclusion}
This paper introduces \liteval and a comprehensive study on literary translation evaluation with human and automatic metrics.

Our study reveals the limitations of error annotation, particularly MQM when comparing top-performing systems. This may be attributed to three factors: (1) the complexity of MQM compared to other schemes; (2) the current MQM framework failing to account for ``errors'' in literary translation intended for certain equivalence in the target language; (3) current leading systems producing fewer errors than previous MT systems, rendering MQM insufficient. While using BWS or experienced professionals yields more plausible assessments, BWS has limitations for further improvements like post-editing when compared to MQM. These results have substantial implications, particularly for determining human parity---a crucial issue in today's LLM landscape. They are especially relevant for professional translators, as naive translation companies might eliminate human translators due to misjudgments based on MQM or SQM evaluations by inexperienced practitioners. While we highlight MQM's limitations in literary MT evaluation, our current study does not propose specific improvements to the existing MQM guidelines. Future work can explore these aspects in depth.

Furthermore, our results indicate that automatic metrics development should focus on aspects such as Style and Terminology. LLMs still need to address the issue of overly literal outputs, and a substantial gap remains between LLM and human quality in literary translation, despite the clear advancements of recent models.

\section*{Limitations}
Several limitations of our current work warrant consideration:

Our study primarily focuses on prompting-based techniques with identical basic prompts across LLMs. While our initial exploration with \gptfo showed no substantial improvement with complex prompts, we acknowledge that LLM-specific prompt optimization could potentially yield better results in literary translation. This aligns with the current understanding that prompting is a delicate process \cite{leiter2024prexme, mizrahi2024state}. 

Other techniques such as fine-tuning with domain knowledge may further enhance LLM performance in this context. Conversely, data contamination may inadvertently improve translation quality for classic works already in the model's training data. However, classics unseen by the model or yet to be translated might exhibit lower translation quality. 

Our focus on paragraph-level translation still does not fully capture the nuances of literary texts. The absence of evaluation on (consecutive) chapters limits our ability to comprehensively evaluate translation quality within a broader context. Although we conducted limited experiments with multi-paragraph samples, resource constraints and the complexity of the task on student evaluators prevented a more extensive exploration in this direction. 

Our study covers four high-resource language pairs, each evaluated by a single evaluator except when computing agreements. We acknowledge that involving multiple evaluators with greater expertise could yield more robust results. To mitigate this limitation within our resource constraints, we implemented a rigorous selection process and provided comprehensive training for our evaluators. However, our dataset doesn't explore LLM performance in low-resource languages---an area that could potentially benefit the most from literary MT.

While we highlight the limitations of MQM in literary MT evaluation, we do not propose specific improvements to the current MQM guidelines. Professional translators agree that MQM is insufficient for literary translation but recognize that enhancing it requires extensive research and exploration. Future work should focus on improving upon MQM to better reflect aspects such as aesthetics, emotion, stylistic features, and the tone of both authors and characters.

\section*{Ethical Considerations}
LLMs in literary translation have sparked widespread public interest and concern. This topic has become a focal point for researchers, translators, and publishers, generating intense debate at various forums including workshops, interviews, and book fairs.\footnote{We list some events below. Workshops:  \url{https://ctt2024.ccl.kuleuven.be/} and \url{https://bohtranslations.com/blog/dual-t-end-of-project-event-recap}. Interviews \& discussions: \url{https://www.goethe.de/ins/gb/en/kul/lue/ail.html},  \url{https://www.buchmesse.de/presse/pressemitteilungen/2023-10-06-kuenstliche-intelligenz}, and \url{https://www.stiftung-genshagen.de/veranstaltungen/veranstaltungen-2024/kuku/default-a39e201a92/}.}  
The increased engagement underscores the need for responsible handling of MT outputs in general.

Low-quality translations can misrepresent or negatively affect an author's original work without their awareness. LLMs may introduce or exacerbate issues of bias and toxicity in translated literature, impacting the audience, particularly children. The growing capabilities of LLMs in translation also threaten professional translators, raising concerns about job displacement and the devaluation of human expertise due to misconceptions about LLM quality. However, the careful use of LLMs or MT models as an auxiliary tool could democratize access to global literary works by improving the accessibility of literature across languages and cultures, especially beneficial for low-resource languages. LLMs could also empower translators, allowing them to focus on the challenging aspects of translation while leaving routine tasks to MT. Additionally, LLMs could serve as potential brainstorming tools, e.g., inspiring more creative wordplay alternatives.

Regarding dataset usage: We adhere to the guidelines specified in Germany's Copyright in Academic Work regulations.\footnote{\url{https://www.bmbf.de/SharedDocs/Publikationen/DE/FS/31580_Urheberrecht_in_der_Wissenschaft_en.pdf}} The dataset usage in this study has been approved by the university ethics committee.

Regarding human evaluation: We obtained consent from all participating evaluators to anonymously disclose their contributions.

\section*{Acknowledgements}
We thank the anonymous reviewers for their thoughtful feedback that greatly improved the text. We also thank all evaluators for their dedication and appreciate the professional input from all translators involved. The NLLG Lab gratefully acknowledges support from the Federal Ministry of Education and Research (BMBF) via the research grant ``Metrics4NLG'' and the German Research Foundation (DFG) via the Heisenberg Grant EG 375/5-1.

\bibliography{latex/acl_latex}

\appendix
\label{sec:appdx}
\section{Corpora comparison}
Table \ref{tab:related_corpus} summarizes the corpora for literary MT (evaluation). We report the annotated dataset and the entire MT parallel corpus separately.  
\clearpage
\thispagestyle{empty}
\begin{landscape}
\centering
\begin{table}\centering
\scriptsize
\begin{tabular}{C{1.5cm}C{2cm}C{1cm}C{1cm}C{1cm}C{1cm}C{1.5cm}C{1.5cm}C{1.2cm}C{2.5cm}C{2.5cm}C{1cm}C{1cm}}\toprule
\multirow{3}{*}{\textbf{Dataset}} &\multicolumn{8}{c}{\textbf{Evaluated Corpus}} &\multicolumn{4}{c}{\multirow{2}{*}{\textbf{Entire Parallel Corpus}}} \\\cmidrule{2-9}
&\multirow{2}{*}{\textbf{Scheme}}&\multicolumn{3}{c}{\textbf{\makecell{\#Evaluated \\segments}}} &\multirow{2}{*}{\textbf{\#systems}} &\multicolumn{2}{c}{\textbf{Source publication}} &\multirow{2}{*}{\textbf{\makecell{Evaluation \\avialability}}} & & & & \\\cmidrule{3-5}\cmidrule{7-8}\cmidrule{10-13}
& &\textbf{\#doc} &\textbf{\#sentences} &\textbf{\#pairs} & &\textbf{Classics} &\textbf{Contemporary} & &\textbf{Source} &\textbf{Target} &\textbf{\#pair} &\textbf{\#doc} \\\midrule
\textbf{BWB} \cite{jiang2022blonde} & \makecell{Span highlight \\ (discourse phenomena)} &80 &2,632 &1 &1 & & \makecell[c]{\checkmark \\(Src: ?) \\(Tgt: ?)} &\checkmark &Chinese &English &1 &196,304 \\
\arrayrulecolor{black!20}\midrule
\textbf{GUOFENG} \cite{xu2022guofeng}& \makecell{Zero Pronoun} &24 &1, 658 &1 &1 & & \makecell[c]{\checkmark \\(Src: ?) \\(Tgt: 2005?-?)}& \checkmark  &Chinese &English &1 &1.5M \\
\arrayrulecolor{black!20}\midrule
\textbf{PAR3} \cite{thai2022exploring} & \makecell{Translation quality: \\ BWS (2) }&300? &unclear &3 &3 & \makecell[c]{\checkmark \\ (Src: 1400-1966?) \\ (Tgt: ?) }& & ?  & (French, Russian, German) + 13 more 
&English &16 &121,385 \\
\arrayrulecolor{black!20}\midrule
\citet{karpinska2023large} &Translation quality; lite-MQM, BWS (2) &1080 &5301 &\textbf{18} &3 & &\makecell[c]{\checkmark \\ (Src: 1992-2021, \\ + 1884 and 1956) \\
(Tgt: 2017-2022)} &\checkmark &English, Polish, Russian, Czech, French, German, Japanese, and Chinese &English, Japanese, Polish &- &- \\
\arrayrulecolor{black!20}\midrule
\textbf{Ours} &\textbf{Translation quality: MQM, SQM, BWS (4)} &\textbf{2197} &\textbf{13,346} &4 &\textbf{10} &\makecell[c]{\textbf{\checkmark} \\ (Src: 1774-1927) \\ (Tgt: 1831-2023)} &\makecell[c]{\textbf{\checkmark} \\ (Src: 2010-2024) \\ (Tgt: 2022-2024)}&\textbf{\checkmark} &\textbf{English, Chinese, German} &\textbf{English, Chinese, German} &\textbf{-} &\textbf{-} \\
\arrayrulecolor{black!100}\bottomrule
\end{tabular}
\caption{Summary of corpora for literary MT (evaluation). We separate the statistics of evaluated samples and the entire MT parallel corpus. We categorize the source works as either classics or contemporary in the source publication columns. We also indicate the time spans for both the source (src) and human translation (tgt) in brackets. "\textbf{?}" denotes an unclear time span. The other 13 languages for PAR3 are Spanish, Czech, Norwegian, Swedish, Portuguese, Italian, Japanese, Bengali, Tamil, Danish, Chinese, Dutch, Hungarian, Polish, Sesotho, and Persian. The \textbf{?} symbol in column evaluation availability indicates that the human evaluation referenced in the paper is not available in the specified GitHub repository.}\label{tab:related_corpus}
\end{table}
\end{landscape}

\section{Corpus details}
Table \ref{tab:meta} summarizes the details of the publications and the corresponding translation versions.
\begin{table*}[!htp]\centering
\scriptsize
\begin{tabular}{clccccc}\toprule
\multirow{2}{*}{\textbf{Pairs}} &\multirow{2}{*}{\textbf{Book}} &\multicolumn{4}{c}{\textbf{Publication year}} & \multirow{2}{*}{\textbf{\#Paragraph}} \\\cmidrule{3-6}
& &\textbf{Source} &\textbf{Translation1} &\textbf{Translation2} &\textbf{Translation3} & \\\midrule
\multirow{4}{*}{\cellcolor[HTML]{d9ead3}\textbf{\enDe}} &\cellcolor[HTML]{d9ead3}Frankenstein &\cellcolor[HTML]{d9ead3}\ul{1818} &\cellcolor[HTML]{d9ead3}\ul{1831} &\cellcolor[HTML]{d9ead3}2018 &\cellcolor[HTML]{d9ead3}2017 & \cellcolor[HTML]{d9ead3}9 \\
&\cellcolor[HTML]{d9ead3}Jane Eyre &\cellcolor[HTML]{d9ead3}\ul{1847} &\cellcolor[HTML]{d9ead3}\ul{1864} &\cellcolor[HTML]{d9ead3}2020 &\cellcolor[HTML]{d9ead3}2008 & \cellcolor[HTML]{d9ead3}13\\
&\cellcolor[HTML]{d9ead3}Wuthering Heights &\cellcolor[HTML]{d9ead3}\ul{1847} &\cellcolor[HTML]{d9ead3}\ul{1941} &\cellcolor[HTML]{d9ead3}\ul{1949} &\cellcolor[HTML]{d9ead3} & \cellcolor[HTML]{d9ead3}8\\
&\cellcolor[HTML]{d9ead3}Birnan Wood &\cellcolor[HTML]{d9ead3}2023 &\cellcolor[HTML]{d9ead3}2023 &\cellcolor[HTML]{d9ead3} &\cellcolor[HTML]{d9ead3} & \cellcolor[HTML]{d9ead3}16\\
\multirow{4}{*}{\cellcolor[HTML]{ead1dc}\textbf{\deen}} &\cellcolor[HTML]{ead1dc}Death in Venice &\cellcolor[HTML]{ead1dc}\ul{1912} &\cellcolor[HTML]{ead1dc}\ul{1925} &\cellcolor[HTML]{ead1dc}2012 &\cellcolor[HTML]{ead1dc}2021 & \cellcolor[HTML]{ead1dc}11\\
&\cellcolor[HTML]{ead1dc}The Metamorphosis &\cellcolor[HTML]{ead1dc}\ul{1915} &\cellcolor[HTML]{ead1dc}2003 &\cellcolor[HTML]{ead1dc}2003 &\cellcolor[HTML]{ead1dc}2014 & \cellcolor[HTML]{ead1dc}13 \\
&\cellcolor[HTML]{ead1dc}Steppenwolf &\cellcolor[HTML]{ead1dc}\ul{1927} &\cellcolor[HTML]{ead1dc}2013 &\cellcolor[HTML]{ead1dc}2012 &\cellcolor[HTML]{ead1dc} & \cellcolor[HTML]{ead1dc}4 \\
&\cellcolor[HTML]{ead1dc}Kairos &\cellcolor[HTML]{ead1dc}2021 &\cellcolor[HTML]{ead1dc}2023 &\cellcolor[HTML]{ead1dc} &\cellcolor[HTML]{ead1dc} & \cellcolor[HTML]{ead1dc}18 \\
\multirow{5}{*}{\cellcolor[HTML]{fff2cc}\textbf{\dezh}} &\cellcolor[HTML]{fff2cc}The sorrows of young werther &\cellcolor[HTML]{fff2cc}\ul{1774} &\cellcolor[HTML]{fff2cc}2018 &\cellcolor[HTML]{fff2cc}1980 &\cellcolor[HTML]{fff2cc} & \cellcolor[HTML]{fff2cc}4\\
&\cellcolor[HTML]{fff2cc}The Metamorphosis &\cellcolor[HTML]{fff2cc}\ul{1915} &\cellcolor[HTML]{fff2cc}2014 &\cellcolor[HTML]{fff2cc}2023 &\cellcolor[HTML]{fff2cc} & \cellcolor[HTML]{fff2cc}2\\
&\cellcolor[HTML]{fff2cc}Demian: Die Geschichte von Emil Sinclairs Jugend &\cellcolor[HTML]{fff2cc}\ul{1921} &\cellcolor[HTML]{fff2cc}2019 &\cellcolor[HTML]{fff2cc}2020 &\cellcolor[HTML]{fff2cc}2014 & \cellcolor[HTML]{fff2cc}16\\
&\cellcolor[HTML]{fff2cc}Steppenwolf &\cellcolor[HTML]{fff2cc}\ul{1927} &\cellcolor[HTML]{fff2cc}2023 &\cellcolor[HTML]{fff2cc}2011 &\cellcolor[HTML]{fff2cc}2022 & \cellcolor[HTML]{fff2cc}13\\
&\cellcolor[HTML]{fff2cc}Heimsuchung &\cellcolor[HTML]{fff2cc}2010 &\cellcolor[HTML]{fff2cc}11.2022 &\cellcolor[HTML]{fff2cc} &\cellcolor[HTML]{fff2cc} & \cellcolor[HTML]{fff2cc}12\\
\multirow{4}{*}{\cellcolor[HTML]{cfe2f3}\textbf{\enzh}} &\cellcolor[HTML]{cfe2f3}Frankenstein &\cellcolor[HTML]{cfe2f3}\ul{1818} &\cellcolor[HTML]{cfe2f3}2010 &\cellcolor[HTML]{cfe2f3}2016 &\cellcolor[HTML]{cfe2f3} & \cellcolor[HTML]{cfe2f3}10\\
&\cellcolor[HTML]{cfe2f3}Jane Eyre &\cellcolor[HTML]{cfe2f3}\ul{1847} &\cellcolor[HTML]{cfe2f3}2012 &\cellcolor[HTML]{cfe2f3}1989 &\cellcolor[HTML]{cfe2f3} & \cellcolor[HTML]{cfe2f3}13 \\
&\cellcolor[HTML]{cfe2f3}Wuthering Heights &\cellcolor[HTML]{cfe2f3}\ul{1847} &\cellcolor[HTML]{cfe2f3}\ul{1956} &\cellcolor[HTML]{cfe2f3}1998 &\cellcolor[HTML]{cfe2f3}2023 & \cellcolor[HTML]{cfe2f3}7\\
&\cellcolor[HTML]{cfe2f3}Chlorine &\cellcolor[HTML]{cfe2f3}2024 &\cellcolor[HTML]{cfe2f3}2024 &\cellcolor[HTML]{cfe2f3} &\cellcolor[HTML]{cfe2f3} & \cellcolor[HTML]{cfe2f3}19 \\
\bottomrule
\end{tabular}
\caption{Meta information for source and translation. \zr{\#Paragraph denotes the number of paragraphs sampled from the book.} Detailed information such as author, translator, and links to the source and translation are contained in the CSV file in the GitHub repository. Classic works, i.e., before year 1980, are \underline{underlined}. }\label{tab:meta}
\end{table*}

\section{Prompt experiments for literary translation}\label{sec:prompt}
We use \gptfo to experiment with four types of prompts. The results are shown in Table \ref{tab:exp_prompt}. 
\begin{table}\centering
\scriptsize
\begin{tabular}{m{0.8cm}m{5.2cm}m{1cm}}\toprule
\textbf{mode} &\textbf{prompt} &\textbf{\%winning} \\\midrule
\textbf{baseline} &Please translate the following literary texts from {src\_lang} to {tgt\_lang}. The texts are as follows: {text} & 27\%\\
\midrule
\textbf{baseline + role} & As a professional literary translator, your task is to translate the provided text with a focus on preserving the original’s literary style, emotional depth, and cultural nuances. Your translation should not be a word-for-word rendering, but rather an adaptation that captures the aesthetic qualities, tone, and creative expression of the source. Ensure the translation reads naturally in the target language, maintaining fluidity and resonance with its literary conventions. 

Pay close attention to conveying the original mood, themes, and subtleties of meaning, while respecting both the integrity of the original work and the target audience’s cultural context. Please translate the following literary texts from {src\_lang} to {tgt\_lang}. The texts are as follows: {text} & -10\%\\
\midrule
\textbf{baseline + multistep error critics} &Please translate the following literary texts from {src\_lang} to {tgt\_lang}. The texts are as follows: {text} 

1. Please translate the source texts in [target language]. 

2. Correct the text with attention to the following error categories: accuracy (including additions, omissions, misnomers, mistranslations such as overly literal translations and temporal inconsistencies, as well as untranslated text), fluency (ensuring consistency, coherence, proper grammar, punctuation, and spelling), style (avoiding awkward phrasing, incorrect register, inconsistencies, and unidiomatic expressions), terminology (using context-appropriate and consistent terms, particularly for culturally specific items and extra-linguistic terms), non-translation (ensuring all text is translated), and adherence to locale conventions. Avoid making any of these errors during the correction process. 

3. Return the corrected translation. & 3\%\\
\midrule
\textbf{baseline + role + multistep error critics} &As a professional literary translator, your task is to translate the provided text with a focus on preserving the original’s literary style, emotional depth, and cultural nuances. Your translation should not be a word-for-word rendering, but rather an adaptation that captures the aesthetic qualities, tone, and creative expression of the source. Ensure the translation reads naturally in the target language, maintaining fluidity and resonance with its literary conventions. Pay close attention to conveying the original mood, themes, and subtleties of meaning, while respecting both the integrity of the original work and the target audience’s cultural context. 

Please translate the following literary texts from {src\_lang} to {tgt\_lang}. The texts are as follows: {text} 

1. First translate the source texts in {tgt\_lang}. 

2. Correct the text with attention to the following error categories: accuracy (including additions, omissions, misnomers, mistranslations such as overly literal translations and temporal inconsistencies, as well as untranslated text), fluency (ensuring consistency, coherence, proper grammar, punctuation, and spelling), style (avoiding awkward phrasing, incorrect register, inconsistencies, and unidiomatic expressions), terminology (using context-appropriate and consistent terms, particularly for culturally specific items and extra-linguistic terms), non-translation (ensuring all text is translated), and adherence to locale conventions. Avoid making any of these errors during the correction process. 

3. Return the corrected translation. & -20\%\\
\bottomrule
\end{tabular}
\caption{Prompt experiments with \gptfo. The comparison is conducted using the \bws scheme. We report the average percentage of language pairs computed with Best-Count minus Worst-count divided by total evaluation count.}\label{tab:exp_prompt}
\end{table}

\section{Human evaluation}
\subsection{Summary of human evaluation schemes}\label{annotation_scheme}
\begin{figure*}
    \centering \includegraphics[width=.89\linewidth]{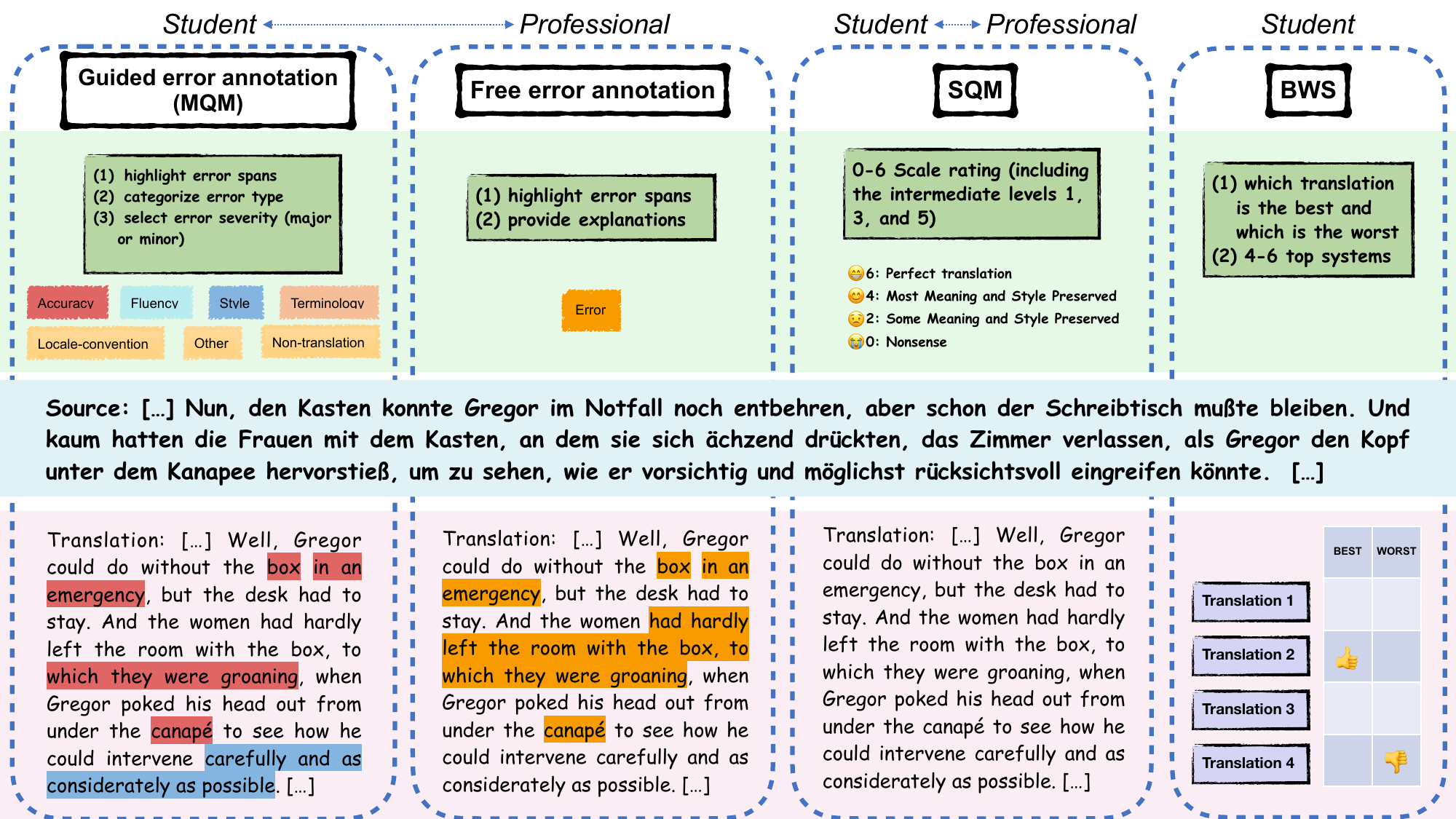}
    \caption{Human evaluation schemes conducted by student evaluators and professional translators with \deen example. MQM, SQM, and BWS represent Multidimensional Quality Metrics, Scalar Quality Metrics, and \bws. MQM is guided by an error categorization guideline and Free annotation permits span highlighting with comments, without specific guidelines.}
    \label{fig:anno_scheme}
\end{figure*}

\zr{Figure \ref{fig:anno_scheme} shows human evaluation schemes with an example.}

\subsection{System ranking}
Table \ref{tab:systranking} shows the system ranking per evaluator and evaluation schemes.
\begin{table*}\centering
\scriptsize
\begin{tabular}{lcccccccc}\toprule
\multirow{2}{*}{\textbf{System}} &\multirow{2}{*}{\textbf{\makecell{Model size \\ (billion)}}} &\multicolumn{4}{c}{Student} &\multicolumn{2}{c}{Professional} \\\cmidrule{3-8}
& &\textbf{MQM } &\textbf{Rank} &\textbf{SQM} &\textbf{Rank} &\textbf{SQM} &\textbf{Rank} \\\midrule
\cellcolor[HTML]{fff2cc}Human &\cellcolor[HTML]{fff2cc}- &\cellcolor[HTML]{fff2cc}-1.3 &\cellcolor[HTML]{fff2cc}\textbf{1} &\cellcolor[HTML]{fff2cc}5.0 &\cellcolor[HTML]{fff2cc}\textbf{1} &\cellcolor[HTML]{fff2cc}5.0 &\cellcolor[HTML]{fff2cc}\textbf{1} \\
\cellcolor[HTML]{b6d7a8}\gtran &\cellcolor[HTML]{b6d7a8}? &\cellcolor[HTML]{b6d7a8}-3.1 &\cellcolor[HTML]{b6d7a8}\textit{3} &\cellcolor[HTML]{b6d7a8}4.0 &\cellcolor[HTML]{b6d7a8}\textit{3} &\cellcolor[HTML]{b6d7a8}2.8 &\cellcolor[HTML]{b6d7a8}\textit{3} \\
\cellcolor[HTML]{b6d7a8}\deepl &\cellcolor[HTML]{b6d7a8}? &\cellcolor[HTML]{b6d7a8}-3.2 &\cellcolor[HTML]{b6d7a8}4 &\cellcolor[HTML]{b6d7a8}3.8 &\cellcolor[HTML]{b6d7a8}4 &\cellcolor[HTML]{b6d7a8}2.2 &\cellcolor[HTML]{b6d7a8}5 \\
\cellcolor[HTML]{fce5cd}\gptfo &\cellcolor[HTML]{fce5cd}>200 &\cellcolor[HTML]{fce5cd}-1.7 &\cellcolor[HTML]{fce5cd}\ul{2} &\cellcolor[HTML]{fce5cd}4.6 &\cellcolor[HTML]{fce5cd}\ul{2} &\cellcolor[HTML]{fce5cd}3.2 &\cellcolor[HTML]{fce5cd}\ul{2} \\
\cellcolor[HTML]{fce5cd}\qwen &\cellcolor[HTML]{fce5cd}7.0 &\cellcolor[HTML]{fce5cd}-8.7 &\cellcolor[HTML]{fce5cd}5 &\cellcolor[HTML]{fce5cd}2.7 &\cellcolor[HTML]{fce5cd}5 &\cellcolor[HTML]{fce5cd} (2.5) &\cellcolor[HTML]{fce5cd}(4) \\
\cellcolor[HTML]{fce5cd}Towerinstruct &\cellcolor[HTML]{fce5cd}7.0 &\cellcolor[HTML]{fce5cd}-9.0 &\cellcolor[HTML]{fce5cd}6 &\cellcolor[HTML]{fce5cd}1.8 &\cellcolor[HTML]{fce5cd}7 &\cellcolor[HTML]{fce5cd} &\cellcolor[HTML]{fce5cd} \\
\cellcolor[HTML]{fce5cd}Llama 3 &\cellcolor[HTML]{fce5cd}8.0 &\cellcolor[HTML]{fce5cd}-12.3 &\cellcolor[HTML]{fce5cd}9 &\cellcolor[HTML]{fce5cd}1.8 &\cellcolor[HTML]{fce5cd}6 &\cellcolor[HTML]{fce5cd} &\cellcolor[HTML]{fce5cd} \\
\cellcolor[HTML]{fce5cd}Gemma &\cellcolor[HTML]{fce5cd}7.0 &\cellcolor[HTML]{fce5cd}-12.6 &\cellcolor[HTML]{fce5cd}10 &\cellcolor[HTML]{fce5cd}1.3 &\cellcolor[HTML]{fce5cd}8 &\cellcolor[HTML]{fce5cd} &\cellcolor[HTML]{fce5cd} \\
\cellcolor[HTML]{f5f5f5}M2M &\cellcolor[HTML]{f5f5f5}3.3 &\cellcolor[HTML]{f5f5f5}-11.2 &\cellcolor[HTML]{f5f5f5}7 &\cellcolor[HTML]{f5f5f5}1.1 &\cellcolor[HTML]{f5f5f5}9 &\cellcolor[HTML]{f5f5f5} &\cellcolor[HTML]{f5f5f5} \\
\cellcolor[HTML]{f5f5f5}NLLB &\cellcolor[HTML]{f5f5f5}1.3 &\cellcolor[HTML]{f5f5f5}-11.9 &\cellcolor[HTML]{f5f5f5}8 &\cellcolor[HTML]{f5f5f5}1.1 &\cellcolor[HTML]{f5f5f5}10 &\cellcolor[HTML]{f5f5f5} &\cellcolor[HTML]{f5f5f5} \\
\bottomrule
\end{tabular}
\caption{System rankings by evaluator and human evaluation scheme. We highlight the top three ranks in \textbf{bold}, \underline{underline}, and \textit{italics}, respectively. Professional translators evaluate \qwen outputs only for \dezh and \enzh pairs, which may result in higher ratings compared to \qwen's performance across all languages.}\label{tab:systranking}
\end{table*}
\subsection{Error distribution}
\begin{figure}
\centering
\begin{subfigure}{0.9\linewidth}
\centering
\includegraphics[width=\linewidth]{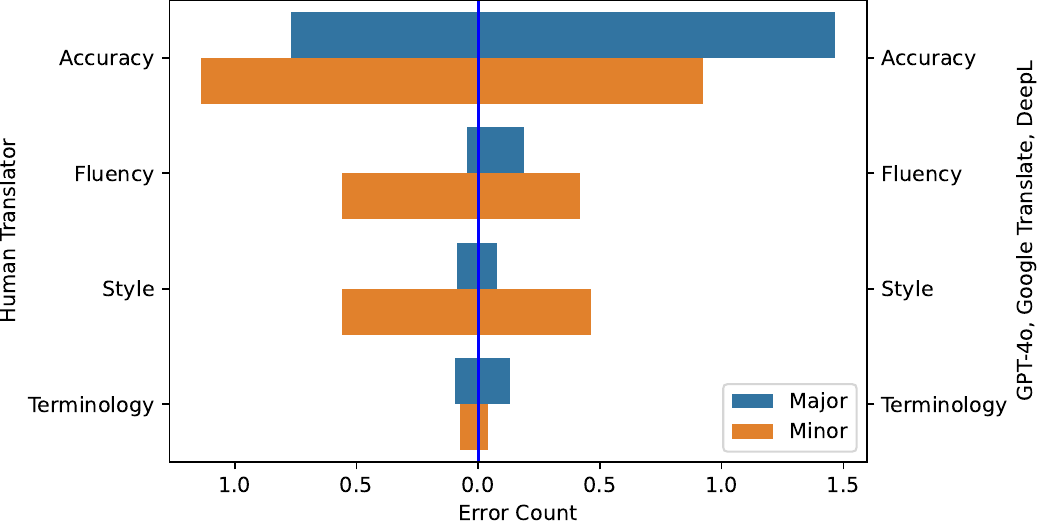} %
\caption{}
\label{fig:errorhumantop3}
\end{subfigure}
\begin{subfigure}{0.9\linewidth}
\centering
\includegraphics[width=\linewidth]{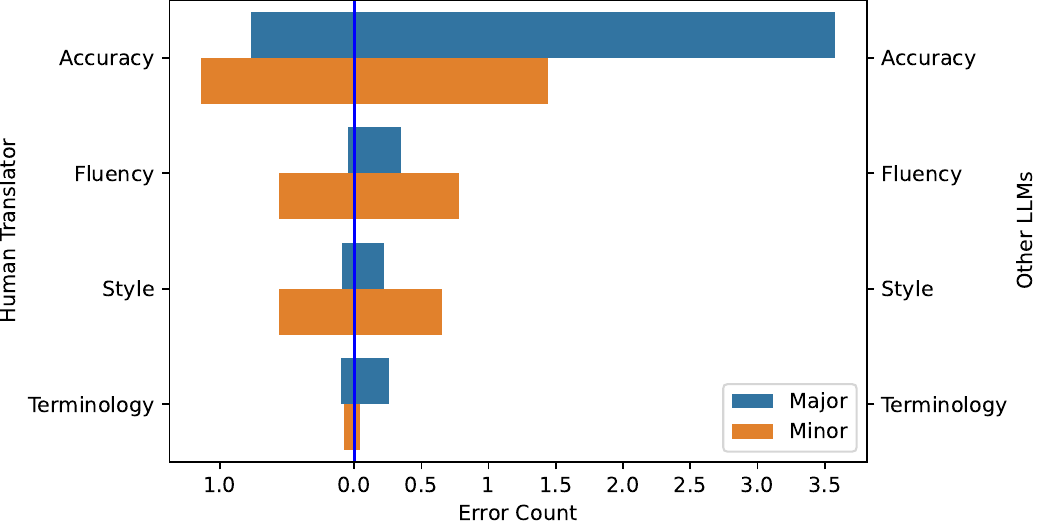} %
\caption{}
\label{fig:errorhumanllms}
\end{subfigure}
\caption{Distribution of mean error count per segment (i.e., the total error count divided by the number of segments) aggregated over error categories: (a) Human Translator vs. \gptfo, \gtran, \deepl. (b) Human Translator vs. other systems.}
\label{fig:errordistibution}
\end{figure}

Figure \ref{fig:errordistibution} shows the distribution of mean error count per segment (i.e., the total error count divided by the number of segments) aggregated over error categories. 

\subsection{Adequacy plot for scheme combination}\label{sec:appdix_adequacy}
\begin{figure}
    \centering
\includegraphics[width=0.9\linewidth]{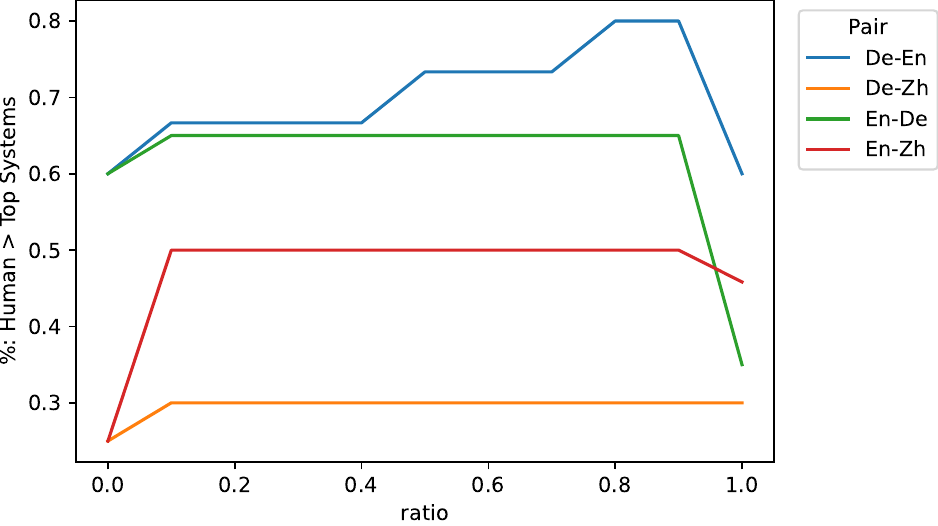}
    \caption{Percentage of segments where human translations are preferred over top systems by different combination ratios $\alpha$, where $\alpha = 0$ corresponds to SQM alone and $\alpha = 1$ corresponds to MQM alone. }
    \label{fig:appdx_adequacy}
\end{figure}
Figure \ref{fig:appdx_adequacy} shows the percentage change in distinguishing human translation from top system outputs by combining student SQM and MQM results. We use min-max scaling \cite{patro2015normalization} to adjust MQM to the same range as SQM (0-6). The combined score is calculated as $(1-\alpha) \times S_\text{MQM} + \alpha \times S_\text{SQM}$, where S represents the score and $\alpha$ represents the ratio between SQM and MQM.

\section{Error annotation details}\label{annotation detail}
\subsection{Total cost}
The four student evaluators each work 60 hours at a rate of 14€ per hour. They are given two months to complete the task. Prior to annotation, the authors and students collaboratively refine the guidelines through several iterations to address any unclear points. Additionally, two professional translators work a total of 20 hours at a rate of approximately 50€ per hour. In total, this comprehensive evaluation project amounts to nearly 4,500€. 

\subsection{MQM Annotation guidelines}\label{guideline}
We present our \zr{annotation guideline in this section and definition of translation errors} in Table \ref{tab:guidelines}. 
\subsubsection{\zr{Annotation guideline}}
\emph{Introduction}

\noindent For each annotation, you will be presented with two paragraphs:
the source paragraph and the translation of the source paragraph. You need to read both paragraphs carefully and finish the following tasks. 

\noindent\emph{Task description}\\
\noindent \textbf{Task 1: MQM error annotation} \\
Step 1: You will be given the definition of translation errors. You need to highlight the spans from the translation containing errors and choose the corresponding error categories. 

Please consider following the order of questions during error annotation:
\begin{itemize}
\setlength\itemsep{-0.3em}
    \item Evaluate whether the sentence is comprehensible. If it is completely incomprehensible, mark it as "non-translation."
    \item Are there any untranslated texts?
    \item Is there any cultural-specific terminology mistranslation or inconsistency?
    \item Any accuracy errors with addition/omission/misnomer or mistranslation especially due to overly literal translation or temporal component?
    \item Are there any style errors?
    \item Are there any fluency errors that cause inconsistency in non-terminology usage and coherence errors?  
\end{itemize}

\noindent Step 2: Rate the severity of errors \\
Major Errors are errors that seriously distort the meaning of the source, in such a way that it becomes completely incomprehensible or that the essence of the message is lost. Another example is errors that may confuse or mislead the reader due to substantial changes in meaning. 

Minor Errors don’t lead to loss of meaning and wouldn’t confuse or mislead the reader but would be noticed, would decrease stylistic quality, fluency, or clarity, or make the content less appealing.
		
You only need to annotate once if the error is repeated several times.

\noindent \textbf{Task 2: SQM assessment}\\
You need to rate the overall translation quality. How do you like the translation, especially in reflecting the stylistic, artistic, and emotional qualities of the original work? Give a score from 0-6.

\subsubsection{Error definition}
\zr{The definition of translation errors is shown in Table \ref{tab:guidelines}. We derive our category definitions primarily from the MQM official website\footnote{\url{https://themqm.org/error-types-2/typology/}} and the study by \citet{wang_2023_findings}. We add two additional categories: \textbf{Non-translation} defined by \citet{freitag_2021_experts} and \textbf{Temporal effect} defined by ourselves.}
\begin{table*}\centering
\scriptsize
\begin{tabular}{m{2cm}m{1.5cm}m{1.8cm}m{8cm}}\toprule
\multicolumn{3}{l}{\textbf{Error Category}} & \textbf{Definition} \\
\toprule
\multirow{7}{*}{Accuracy} &\multicolumn{2}{l}{Addition} &Error occuring in the target content that includes content not present in the source. \\
\cmidrule{2-4}
&\multicolumn{2}{l}{Omission} &Error where content present in the source is missing in the target. \\
\cmidrule{2-4}
&\multicolumn{2}{l}{Misnomer} &The target text is more/less specific than the raw \\
\cmidrule{2-4}
&\multirow{3}{*}{Mistranslation} & General &Translation does not accurately represent the source. Wrong translation in the target language. E.g., the wrong word is used in the target language; mistranslation of a false friend (word or expression that has a similar form in the source and target languages, but a different meaning); \\
\cmidrule{3-4}
& &Overly literal &Direct translation of idioms, sentences, and structures; \\
\cmidrule{3-4}
& &Temporal effect &Historical linguistic elements such as spelling, semantics, and morphology that affect the accuracy of translation. \\
\cmidrule{2-4}
&\multicolumn{2}{l}{Untranslated} &Error occuring when a text segment that was intended for translation is omitted in the target content. \\
\midrule
\multirow{4}{*}{Fluency} &\multicolumn{2}{l}{Punctuation/Spelling} &Punctuation marks missing or used in the wrong way/Issues related to the spelling of words (Including those of capitalization hyphenated words and use of as risk for censored swear words.) \\
\cmidrule{2-4}
&\multicolumn{2}{l}{Grammar} &Issues related to the grammar or syntax of the text. other than spelling and orthography (especially inconsistency of the tenses and conditionals) \\
\cmidrule{2-4}
&\multicolumn{2}{l}{Inconsistency} &The text shows internal inconsistency \\
\cmidrule{2-4}
&\multicolumn{2}{l}{coherence} &The text is not semantically clear, logical, and consistent, and, therefore, it cannot be understood by the reader; E.g., - Unclear reference: Relative pronouns or other referential mechanisms unclear in their reference. \\
\midrule
\multirow{4}{*}{Style} &\multicolumn{2}{l}{Awkwardness} &Style involving excessive wordiness or overly embedded clauses, often due to inappropriate retention of source text style in the target text. \\
\cmidrule{2-4}
&\multicolumn{2}{l}{Register} &Characteristic of text that uses a level of formality higher or lower than required by the specifications or general language conventions. \\
\cmidrule{2-4}
&\multicolumn{2}{l}{Inconsistent} &Style is inconsistent within a text \\
\cmidrule{2-4}
&\multicolumn{2}{l}{Unidiomatic} &Style that is grammatical, but unnatural. \\
\midrule
\multirow{2}{*}{Terminology} &\multicolumn{2}{l}{Mistranslation} &A genre-specific or cultural-specific terminology is wrongly translated. Use of term that it is not the term a domain expert would use or because it gives rise to a conceptual mismatch. \\
\cmidrule{2-4}
&Inconsistent & &Terminology is used in a consistent manner within the text \\
\midrule
\multirow{2}{*}{\makecell{Locale Convention \\ (non-cultrual)}} &\multicolumn{2}{l}{Location Format} &Using the wrong format for address, name etc. \\
\cmidrule{2-4}
&\multicolumn{2}{l}{Number Format} &The translated date, time, currency, and telephone use formats inappropriate for its locale. \\
\midrule
\multicolumn{3}{l}{Others} &Other issues that haven‘t been included in this list. \\
\midrule
\multicolumn{3}{l}{Non-translation} &Non-translation error that can be used to tag an entire sentence which is too badly garbled to permit reliable identification of individual errors. \\
\bottomrule
\end{tabular}
\caption{MQM \zr{error definition}}\label{tab:guidelines}
\end{table*}

\subsubsection{Screenshot of error annotation platform}
\begin{figure*}
    \centering
\includegraphics[width=0.65\linewidth, angle=-90]{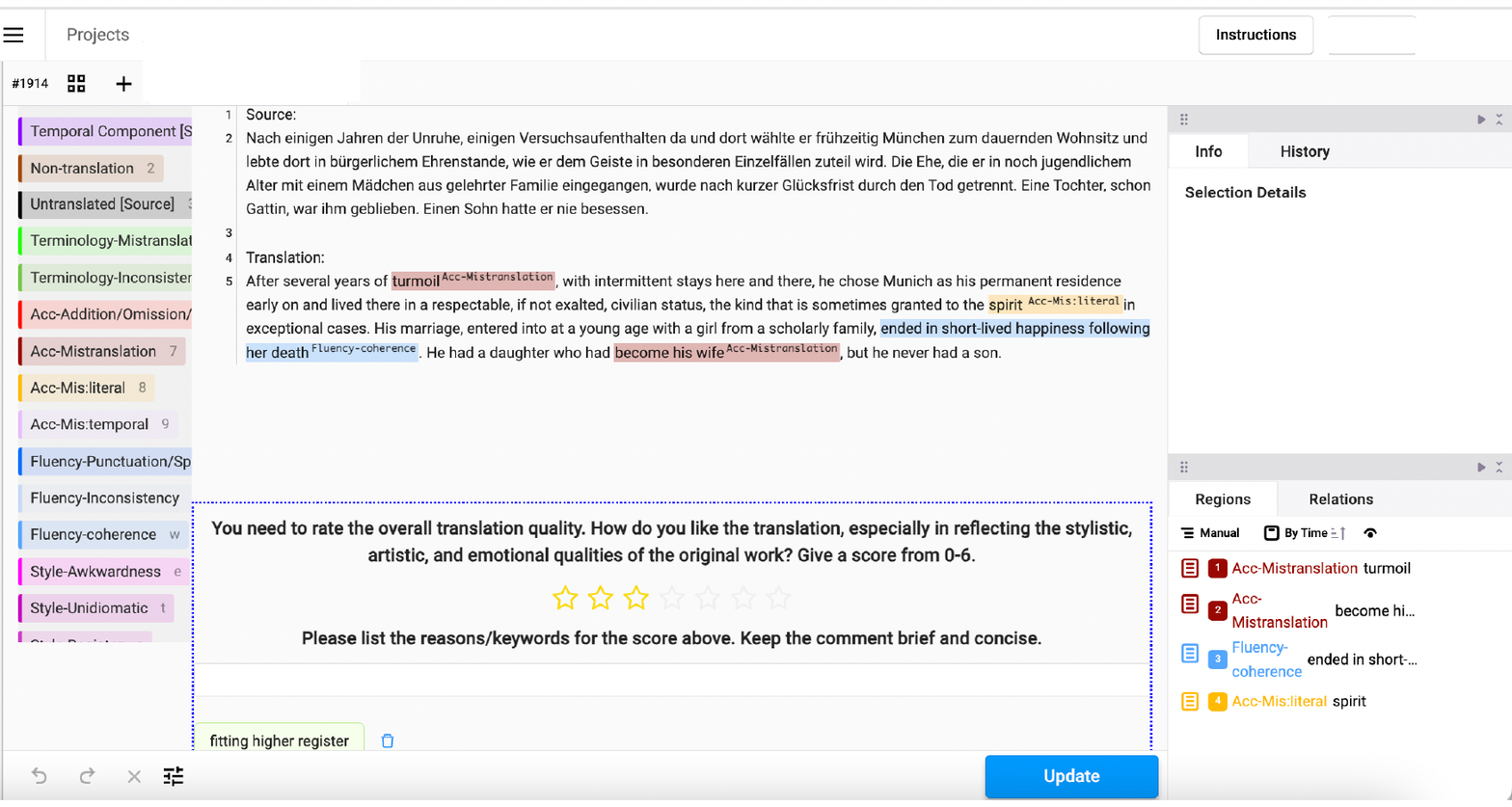}
    \vspace*{-15mm}\caption{Screenshot of error annotation interface}
    \label{fig:screenshot}
\end{figure*}
\zr{Figure \ref{fig:screenshot} demonstrates the screenshot of the error annotation platform.}

\subsection{Examples of human evaluation results}\label{sec:example}
\subsubsection{\deen}
\begin{figure*}
    \centering    \includegraphics[width=1\linewidth]{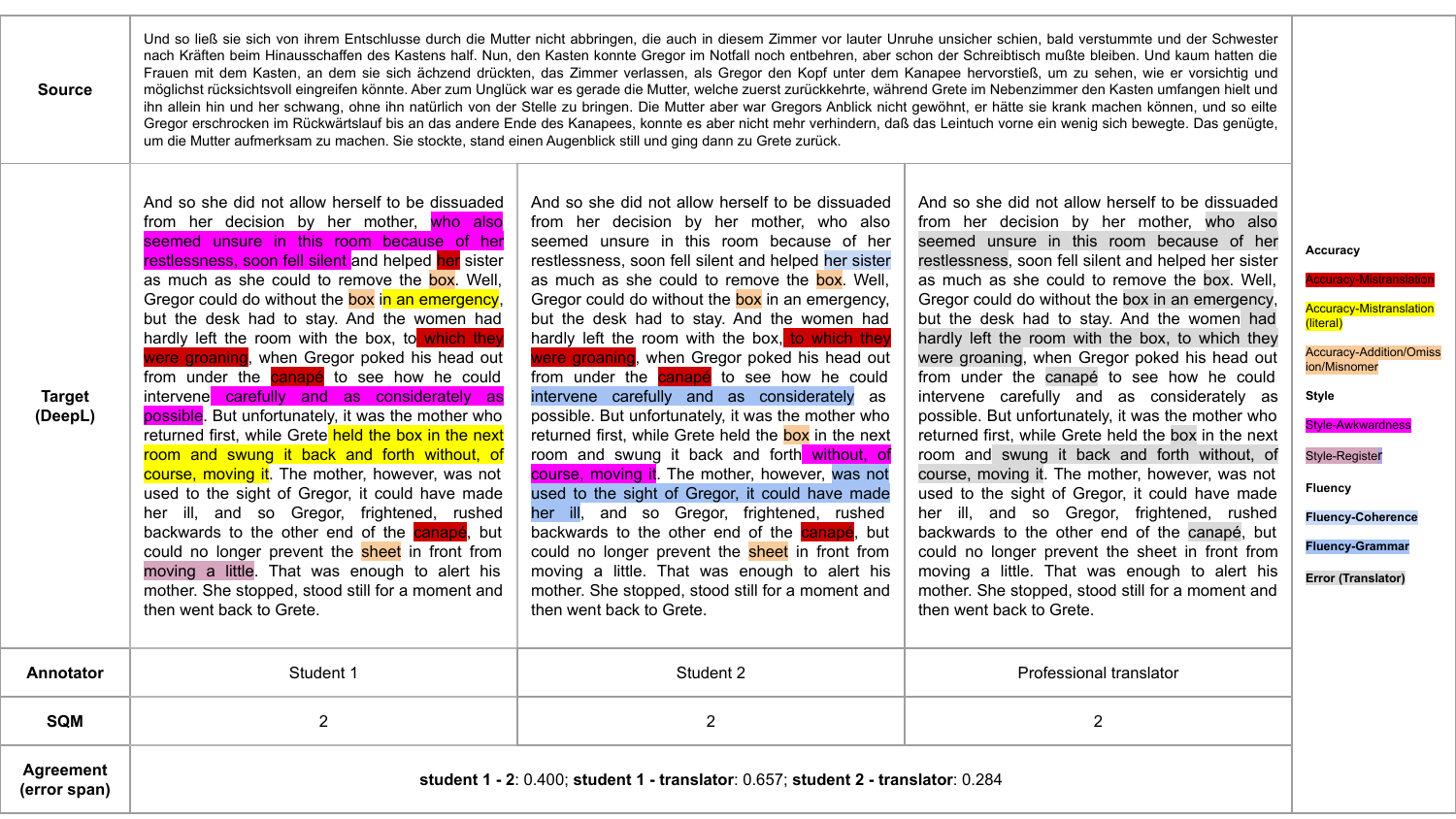}
    \caption{Annotation example for \deen}
    \label{fig:exp_deen_good}
\end{figure*}

Figure \ref{fig:exp_deen_good} shows examples of SQM and error annotations by student and professional evaluators.

\subsection{Example of human translation labelled as ``error'' according to student MQM}
\zr{Table \ref{tab:mis_label} shows examples of MQM errors labeled by student evaluators due to cultural adaptation in the translation.}
\begin{table}\centering
\scriptsize
\begin{tabular}{p{0.09\linewidth}p{0.15\linewidth}p{0.2\linewidth}p{0.35\linewidth}}\toprule
pair &Source &Human Translation &Interpretation \\\midrule
\enzh & ... a wilderness of crumbling (griffins) and shameless little boys ... & ... \begin{CJK*}{UTF8}{gbsn}许多残破的(怪兽)和不知羞的小男孩 \end{CJK*}... & The text in brackets is marked as a misnomer (a subcategory of accuracy error) by the student evaluator. While the source text mentions griffins, the translation simply uses "monster," since in the target culture, griffins as mythical creatures have no cultural significance. \\
\midrule
\dezh &Als (Lateinschüler) und Herrensöhnchen ... & \begin{CJK*}{UTF8}{gbsn}我出身良好，读(高级中学)\end{CJK*} ... &The text in brackets is marked as a mistranslation (a subcategory of accuracy error) by the student evaluator. Though the source text mentions "Latin school," the translation uses "prestigious school" because Latin schools don't exist in the target culture and therefore lack the same connotation of academic prestige. \\
\midrule
\deen &... ein (dreister Schlager) in unverständlichem Dialekt ... & ... a ("big number") in incomprehensible dialect ... &The text in brackets is marked as a mistranslation (a subcategory of accuracy error) by the student evaluator, which is a culturally adapted translation. \\
\bottomrule
\end{tabular}
\caption{Mislabelled MQM error due to cultural adaptation}\label{tab:mis_label}
\end{table}

\section{Automatic metrics}
\subsection{Detail of automatic metrics}\label{sec:autometric}
Table \ref{tab:metric_sum} summarizes the characteristics of automatic metrics. For cost reasons, we use GPT-4o-mini instead of GPT-4 used in the original paper. This may induce a slight performance reduction. But compared to the cost for GPT-4 (on average 0.12€ per segment, approximately 250€ for complete evaluation per prompt), GPT-4o-mini is over 200 times cheaper (1.2€ for full evaluation per prompt).
\begin{table}[H]\centering
\scriptsize
\begin{tabular}{lcccc}\toprule
&\textbf{aspect} &\textbf{finetuned} &\textbf{base model} &\textbf{\#parameter} \\\midrule
\textbf{Prometheus 2} & & &prometheus2 &7 billion \\
\textbf{\cometxl} & &X &XLMR-XL &3.5 billion \\
\textbf{\cometxxl} & &X &XLMR-XXL &10 billion \\
\textbf{\gemba} &X & &GPT-4o mini &? \\
\bottomrule
\end{tabular}
\caption{Summary of characteristics for automatic metrics. \textit{Aspect}  indicates whether the metric support evaluation per aspect such as error categories and \#parameter indicates the size of the base model.}\label{tab:metric_sum}
\end{table}

\subsection{\gemba detail}\label{sec:gemba_detail}

\subsubsection{\gemba prompt templates}
Table \ref{tab:gembatemplate} showcases the prompt templates. 
\begin{table*}
\scriptsize
\begin{tabular}{m{2cm}m{6.2cm}m{6.2cm}}\toprule
&\gemba (Original) &\gemba (Literary) \\\midrule
system &You are an annotator for the quality of machine translation. Your task is to identify errors and assess the quality of the translation. & As a literary translation critic, your role is to identify errors and evaluate the translation's quality. Focus on the subtleties of literary style, emotional impact, and creative expression. An excellent translation captures the original work's aesthetic qualities, tone, and cultural nuances, rather than adhering to a word-for-word approach. Translations that are excessively literal and fail to adapt to the target language's literary conventions and natural flow should be critiqued accordingly. \\
\midrule
prompt & 
\multicolumn{2}{m{13cm}}{Based on the source segment and machine translation surrounded with triple backticks, identify error types in the translation and classify them. The categories of errors are: accuracy (addition, omission, misnomer, mistranslation [including too-literal translation and temporal effect], untranslated text), fluency (inconsistency, coherence, grammar, punctuation, spelling), style (awkward, register, inconsistent, unidiomatic), terminology (inappropriate for context, inconsistent use, please pay attention to cultural specific items and extra-linguistic terms), non-translation, other, locale convention, or no error.  
Each error is classified as one of three categories: critical, major, and minor. Critical errors inhibit comprehension of the text. Major errors disrupt the flow, but what the text is trying to say is still understandable. Minor errors are technically errors, but do not disrupt the flow or hinder comprehension.  }\\
\midrule
few shot example 1 & \enDe: \{ ``source\_lang'': ``English'', ``source\_seg'': ``I do apologise about this, we must gain permission from the account holder to discuss an order with another person, I apologise if this was done previously, however, I would not be able to discuss this with yourself without the account holders permission.'', ``target\_lang'': ``German'', ``target\_seg'': ``Ich entschuldige mich dafür, wir müssen die Erlaubnis einholen, um eine Bestellung mit einer anderen Person zu besprechen. Ich entschuldige mich, falls dies zuvor geschehen wäre, aber ohne die Erlaubnis des Kontoinhabers wäre ich nicht in der Lage, dies mit dir involvement.'', ``answer'': Critical: no-error Major: accuracy/mistranslation - ``involvement'' accuracy/omission - ``the account holder'' Minor: fluency/grammar - ``wäre'' style/register - ``dir'' \}& \enDe: \{ ``source\_lang'': ``English'', ``source\_seg'': ``At intervals, while turning over the leaves of my book, I studied the aspect of that winter afternoon.'', ``target\_lang'': ``German'', ``target\_seg'': ``Von Zeit zu Zeit, während ich in meinem Buch blätterte, studierte ich das Aussehen dieses Winternachmittags.'', ``answer'': Critical: accuracy/mistranslation (Too-literal) - ``studierte'' Major: accuracy/omission - ``das Aussehen'' Minor: no-error \} \\
few shot example 2 & \begin{CJK*}{UTF8}{gbsn} \zhen: \{``source\_lang'': ``Chinese'', ``source\_seg'': ``大众点评乌鲁木齐家居卖场频道为您提供高铁居然之家地址，电话，营业时间等最新商户信息，找装修公司，就上大众点评'', ``target\_lang'': ``English'', ``target\_seg'': ``Urumqi Home Furnishing Store Channel provides you with the latest business information such as the address, telephone number, business hours, etc., of high-speed rail, and find a decoration company, and go to the reviews.'', ``answer'': Critical: accuracy/addition - ``of high-speed rail'' Major: accuracy/mistranslation - ``go to the reviews'' Minor: style/awkward - ``etc.,'' \} \end{CJK*}
& \begin{CJK*}{UTF8}{gbsn}  \zhen: \{
            ``source\_lang'': ``Chinese'',
            ``source\_seg'': ``太阳他有脚啊，轻轻悄悄地挪移了。'',
            ``target\_lang'': ``English'',
            ``target\_seg'': ``The sun he has feet, ah, gently and quietly moved.'',
            ``answer'': Critical:
style/awkward - ``ah'' 
Major:  
fluency/grammar - ``gently and quietly moved``
Minor:
accuracy/mistranslation (Too-literal) - ``he has feet''
        \} \end{CJK*}\\
\bottomrule
\end{tabular}
\caption{\gemba prompt templates.}\label{tab:gembatemplate}
\end{table*}

\subsubsection{Consistency analysis of \gemba (Literary) under different temperature}
\zr{Table \ref{tab:consistency} shows the consistency of \gemba (Literary) under different temperature. We evaluate the consistency of \gemba (Literary) on the full \deen dataset under varying temperatures, each temperature querying three times. From Table \ref{tab:consistency}, we observe that even at temperature 0 (the deterministic setting), the resulting score shows slight variation while maintaining high overall consistency. The correlation coefficient between different queries decreases from 0.910 at temperature 0.0 to 0.856 at temperature 1.0. Similarly, the proportion of cases with small score differences ($\delta \leq 1$) drops from 78.8\% to 73.3\%, while larger differences ($\delta > 5$) increase from 4.3\% to 7.5\% when temperature increases. These results demonstrate that the scores remain robust and consistent at temperature 0 and across low temperatures but variability increases with higher temperatures. }
\begin{table}[H]
\centering
\scriptsize
\begin{tabular}{cccccc}\toprule
\multirow{2}{*}{\textbf{temperature}} &\multirow{2}{*}{\textbf{corr}} &\multicolumn{3}{c}{\textbf{score difference $\delta$}} \\\cmidrule{3-5}
& &\textbf{$\delta \leq 1 $ } &\textbf{ $ 1 < \delta \leq 5 $} &\textbf{$\delta > 5 $ } \\\midrule
0.0 &0.910 &78.8\% &16.9\% &4.3\% \\
0.1 &0.902 &76.9\% &16.9\% &6.2\% \\
0.3 &0.882 &74.7\% &18.7\% &6.6\% \\
0.5 &0.887 &71.9\% &22.6\% &5.5\% \\
0.7 &0.874 &72.2\% &20.5\% &7.3\% \\
1.0 &0.856 &73.3\% &19.2\% &7.5\% \\
\bottomrule
\end{tabular}
\caption{\zr{Consistency of \gemba scores across different temperatures. We query three times at each temperature. Corr calculates the mean rank correlation between queries using Spearman correlation. The score difference shows the percentage of score changes under different thresholds. $\delta = 1$ represents one minor error in difference, while $\delta = 5$ represents one major error.}}\label{tab:consistency}
\end{table}

\subsection{Prompt for Prometheus 2}
\begin{table}[H]\centering
\scriptsize
\begin{tabular}{p{0.19\linewidth}p{0.7\linewidth}}\toprule
Component &prompt \\\midrule
ABS\_SYSTEM PROMPT &As a literary translation critic, your role is to identify errors and evaluate the translation’s quality. Focus on the subtleties of literary style, emotional impact, and creative expression. An excellent translation captures the original work’s aesthetic qualities, tone, and cultural nuances, rather than adhering to a word-for-word approach. Translations that are excessively literal and fail to adapt to the target language’s literary conventions and natural flow should be critiqued accordingly. \\
\midrule
\multirow{4}{*}{rubric\_data} & \textbf{``criteria''}: ``What is the overall quality of the given literary translation based on the source texts?'' \\
& \textbf{``score0\_description''}: ``Nonsense: Nearly all information is lost between the translation and source. Grammar and style are irrelevant.''\\
&\textbf{``score2\_description''}: ``Some Meaning and Style Preserved: The translation preserves some of the meaning and style of the source but misses significant parts. The narrative is hard to follow due to fundamental errors. Grammar may be poor. Style may be inconsistent.'' \\
&\textbf{``score4\_description''}: ``Most Meaning and Style Preserved and Few Grammar Mistakes: The translation retains most of the meaning and style of the source. This may contain some grammar mistakes or minor style and contextual inconsistencies.'' \\
&\textbf{``score6\_description''}: ``Perfect Meaning and Style Preserved: The meaning and style of the translation are completely consistent with the source and the surrounding context.''\\
\bottomrule
\end{tabular}
\caption{Prompts for Prometheus 2}\label{tab:Prometheus}
\end{table}
\subsection{Correlation between SQM and automatic metrics}
\begin{figure}[H]
    \centering
    \includegraphics[width=0.95\linewidth]{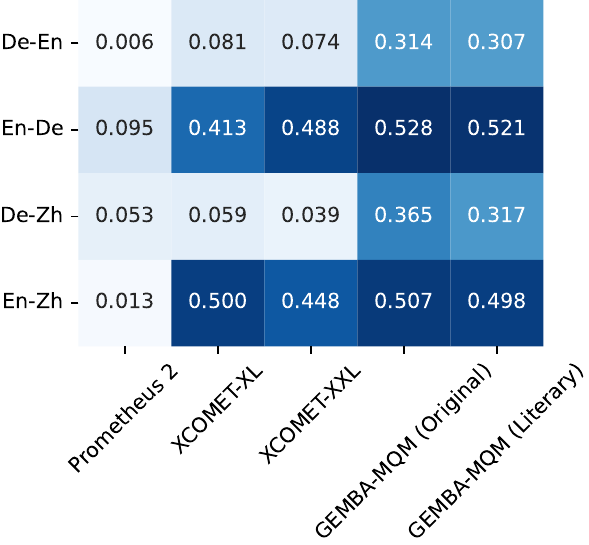}
    \caption{Segment level correlation measured by Kendall’s tau between human SQM and automatic metrics.}
    \label{fig:sqm_corr}
\end{figure}

Figure \ref{fig:sqm_corr} illustrates segment-level correlation between human SQM and automatic metrics.
\subsection{Human vs. LLMs for metrics.}\label{sec:preference}
Table \ref{tab:humanvsllm_metric} shows the percentage of segments where human translations are preferred over machine translations per evaluation schemes and automatic metrics. 
\clearpage
\thispagestyle{empty}
\begin{landscape}
\centering
\begin{table}\centering
\scriptsize
\begin{tabular}{cccccccccccccccccc}\toprule
\multirow{3}{*}{\textbf{pair}} &\multicolumn{8}{c}{\textbf{\makecell{Human > Top Systems \\ (GPT-4o, DeepL, Google Translate, Qwen 2)}}} &\textbf{} &\multicolumn{8}{c}{\textbf{\makecell{Human > Other Systems \\ (excluding GPT-4o, DeepL, Google Translate)}}} \\\cmidrule{2-9}\cmidrule{11-18}
&\multicolumn{2}{c}{\textbf{Student}} &\multirow{2}{*}{\textbf{\makecell[c]{Prometheus \\2}}}  &\multicolumn{2}{c}{\textbf{XCOMET}} & &\multicolumn{2}{c}{\textbf{\gemba }} & &\multicolumn{2}{c}{\textbf{Student}} &\multirow{2}{*}{\textbf{\makecell[c]{Prometheus \\2}}} &\multicolumn{2}{c}{\textbf{XCOMET}} & &\multicolumn{2}{c}{\textbf{\gemba }} \\\cmidrule{2-3}\cmidrule{5-6}\cmidrule{8-9}\cmidrule{11-12}\cmidrule{14-15}\cmidrule{17-18}
&\textbf{MQM} &\textbf{SQM}  & &\textbf{XL} &\textbf{XXL} & &\textbf{Original} &\textbf{Literary} & &\textbf{MQM} &\textbf{SQM} & &\textbf{XL} &\textbf{XXL} & &\textbf{Original} &\textbf{Literary} \\\midrule
\textbf{\deen} &60.0\% &60.0\% &6.7\% &6.7\% &13.3\% & &6.7\% &0.0\% & &86.7\% &80.0\% &6.7\% &20.0\% &33.3\% & &26.7\% &33.3\% \\
\textbf{\enDe} &35.0\% &60.0\% &10.0\% &5.0\% &20.0\% & &0.0\% &10.0\% & &95.0\% &100.0\% &10.0\% &70.0\% &75.0\% & &70.0\% &60.0\% \\
\textbf{\dezh} &30.0\% &25.0\% &15.0\% &30.0\% &10.0\% & &15.0\% &20.0\% & &95.0\% &90.0\% &25.0\% &30.0\% &45.0\% & &45.0\% &50.0\% \\
\textbf{\enzh} &45.8\% &25.0\% &8.3\% &20.8\% &41.7\% & &0.0\% &8.3\% & &87.5\% &79.2\% &4.2\% &70.8\% &75.0\% & &41.7\% &50.0\% \\
\hdashline
\textbf{Mean} &42.7\% &42.5\%  &10.0\% &15.6\% &21.3\% & &5.4\% &9.6\% & &91.0\% &87.3\% &11.5\% &47.7\% &57.1\% & &45.8\% &48.3\% \\
\bottomrule
\end{tabular}
\caption{\label{tab:humanvsllm_metric}
   Percentage of segments where human translations are preferred over machine translations per human evaluation schemes and automatic metrics.}
\end{table}
\end{landscape}

\zr{Figure \ref{fig:adequacy} is based on Table \ref{tab:humanvsllm} and Table \ref{tab:humanvsllm_metric} (appendix). For the percentage computation of scheme professional error annotation, we first convert error span annotation counts to numerical scores using the formula $\text{Score} = |\text{Good solution}| - |\text{Error}|$, then rank these scores to compute the percentage. The percentage of human > top systems averaged over three language pairs is 62.8\%, with \deen at 75.0\%, \enDe at 33.3\%, and \enzh 80.0\%, respectively. }

\zr{The professional BWS statistics shown in Figure \ref{fig:adequacy} are based on the results from \enzh, with 14 samples evaluated by the \enzh translator.}

\subsection{Computation details for diversity measure}\label{sec:formula_diversity}
To assess a system's lexical diversity compared to others, we calculate the pairwise lexical overlap between translations from different systems. For each paragraph, we compute the lexical overlap between the current system's translation and those of all other systems. This overlap quantifies the vocabulary shared between the current system and others. We then average these overlaps for all paragraphs, resulting in an overall lexical overlap score for the system. A lower average overlap indicates that the system uses a more diverse and distinct vocabulary, while a higher score suggests greater similarity in word choice with other systems.

The mean lexical overlap for system \( j \) across all paragraphs is given by:

\[
\text{Avg.overlap}_j = \frac{1}{N} \sum_{i=1}^{N} \left( \frac{1}{M-1} \sum_{\substack{k=1 \\ k \neq j}}^{M} S(T_{ij}, T_{ik}) \right)
\]

\noindent where  \( N \) is the total number of paragraphs. \( M \) is the total number of systems.
 \( T_{ij} \) is the translation of paragraph \( i \) by system \( j \).
 $S(T_{ij}, T_{ik})$ is the function, i.e., BLEU by SACREBLEU \cite{post-2018-sacrebleu}, to compute lexical overlap between the translation from system \( j \) and system \( k \) for paragraph \( i \).

\subsection{Comparison of classic and contemporary works}
\begin{figure}
    \centering
\includegraphics[width=0.8\linewidth]{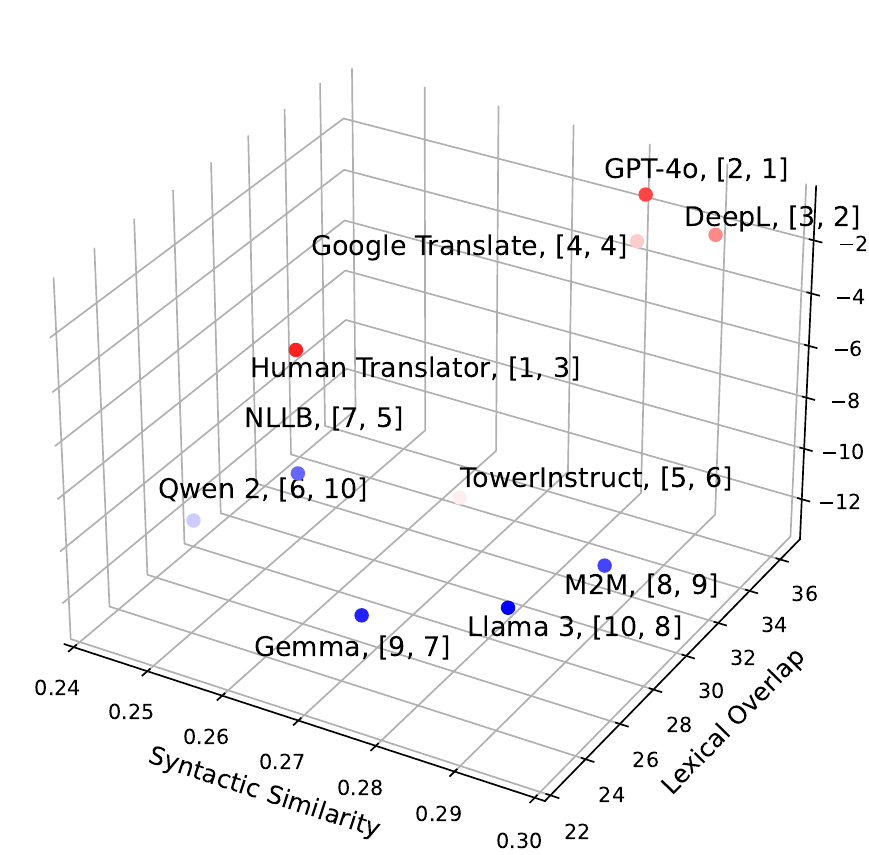}{}%
    \caption{(Contemporary works after 2000) Scatter plots of human MQM (z-axis) per average syntactic similarity between source and system translation and average lexical overlap of the current system to all other systems. The rank of systems is marked as [rank per human MQM, rank per \gemba (Literary)].}\label{fig:syntacticsym_recent}
\end{figure}
\zr{Figure \ref{fig:syntacticsym_recent}
shows the scatter plots of human MQM (z-axis) per average syntactic similarity between source and system
translation and average lexical overlap of the current system to all other systems for all contemporary works after 2000. Human translation still holds the lowest lexical overlap (19.6) than other systems except for Qwen 2 (19.5). The syntactic similarity of human translation (0.267) ranks third slightly higher than Qwen2 (0.252) and NLLB (0.253).}

\begin{figure}
    \centering
    \includegraphics[width=0.9\linewidth]{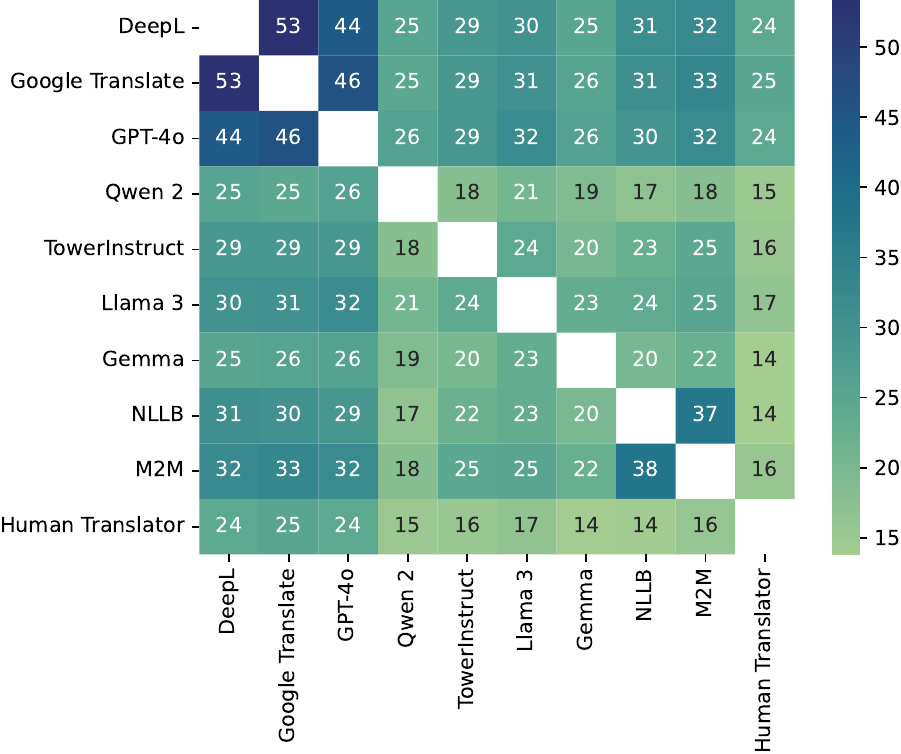}
    \caption{Heatmap for pairwise lexical overlap between translation systems (calculation based on the entire \liteval).}
    \label{fig:overlap}
\end{figure}

\zr{Figure \ref{fig:overlap} shows the heatmap of pairwise lexical overlap between systems based on the entire \liteval. The results reveal exceptionally high lexical overlap among \gtran, \gptfo, and DeepL, while NLLB and M2M (all from META AI) also show notably high overlap. This pattern may stem from shared training corpora or training data distillation from other models.} 

\subsection{Comparison of different human translation versions}
\begin{figure}[H]
    \centering
\includegraphics[width=0.99\linewidth]{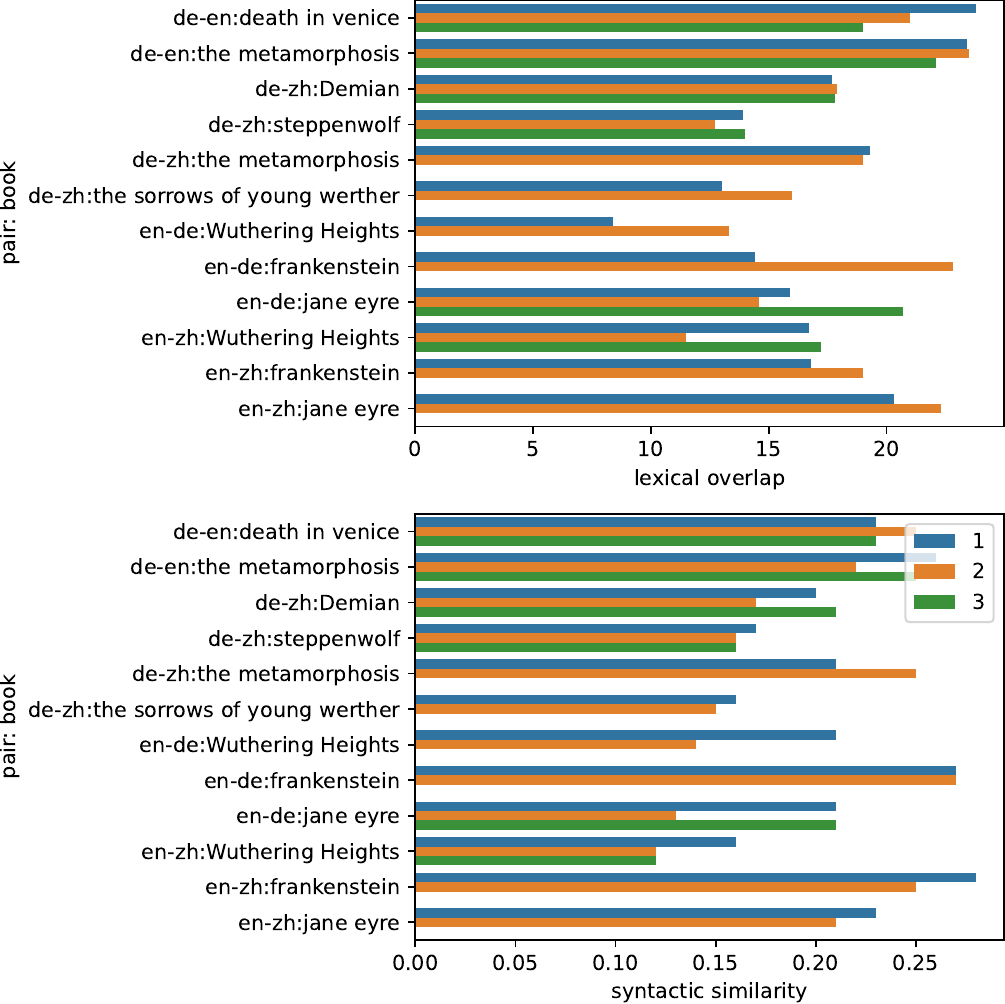}
    \caption{Comparison of lexical overlap and syntactic similarity for different human translation versions.}\label{fig:translation_version}
\end{figure}

\zr{Figure \ref{fig:translation_version} compares different human translation versions over time of the same source per lexical overlap (upper) and syntactic similarity (lower). Versions 1-3 represent translations from the oldest to newest versions in our study. Our calculations are based on paragraphs from our \liteval. For lexical overlap with other LLM systems (upper), 8 out of 12 books show that the newest translations show substantially higher overlap with other system translations (e.g., Wuthering Heights, Frankenstein, and Jane Eyre) compared to previous versions. However, some older translations (e.g., Wuthering Heights \enzh and Death in Venice \deen) also show high overlap, likely because they are part of the training data of the translation systems. Regarding syntactic similarity between source and translation (lower), we observe no consistent pattern across versions, with some translations like the Metamorphosis \dezh showing higher syntactic similarity while others remain lower or comparable.}
 
\subsection{Example of system outputs with human and metric scores, syntactic similarity, and lexical overlap}
Table \ref{tab:syntax_example1} and \ref{tab:syntax_example2} demonstrate a full example including all system outputs with evaluation from human and automatic metrics, syntactic similarity to the source, and lexical overlap to all other systems. 
\begin{table*}
  \centering
\tiny
\begin{tabular}{C{1cm}m{7cm}C{1cm}C{1cm}C{1cm}C{1cm}C{1cm}}\toprule
\multicolumn{7}{c}{\textbf{Source}} \\
\multicolumn{7}{m{15.5cm}}{Aus diesem Traum erwachte der Heimgesuchte entnervt, zerrüttet und kraftlos dem Dämon verfallen. Er scheute nicht mehr die beobachtenden Blicke der Menschen; ob er sich ihrem Verdacht aussetze, kümmerte ihn nicht. Auch flohen sie ja, reisten ab; zahlreiche Strandhütten standen leer, die Besetzung des Speisesaals wies größere Lücken auf, und in der Stadt sah man selten noch einen Fremden. Die Wahrheit schien durchgesickert, die Panik, trotz zähen Zusammenhaltens der Interessenten, nicht länger hintanzuhalten. Aber die Frau im Perlenschmuck blieb mit den Ihren, sei es, weil die Gerüchte nicht zu ihr drangen, oder weil sie zu stolz und furchtlos war, um ihnen zu weichen: Tadzio blieb; und jenem, in seiner Umfangenheit, war es zuweilen, als könne Flucht und Tod alles störende Leben in der Runde entfernen und er allein mit dem Schönen auf dieser Insel zurückbleiben,—ja, wenn vormittags am Meere sein Blick schwer, unverantwortlich, unverwandt auf dem Begehrten ruhte, wenn er bei sinkendem Tage durch Gassen, in denen verheimlichterweise das ekle Sterben umging, ihm unwürdig nachfolgte, so schien das Ungeheuerliche ihm aussichtsreich und hinfällig das Sittengesetz.} \\
\midrule
\textbf{Model} &\textbf{Translation} &\textbf{Syntactic similarity (Src-Tgt)} &\textbf{Lexical overlap} &\textbf{Human MQM} &\textbf{SQM} &\textbf{\gemba (Literary)} \\
\midrule
Human Translator 1 &When he awakened from the affliction of this dream he was unnerved, shattered, and hopelessly under the power of the demon. He no longer avoided the inquisitive glances of other people; he did not care if he was exciting their suspicions. And as a matter of fact they were fleeing, travelling elsewhere. Numerous bathing houses stood empty, the occupants of the dining-hall became more and more scattered, and in the city now one rarely saw a foreigner. The truth seemed to have leaked out; the panic, despite the reticence of those whose interests were involved, seemed no longer avoidable. But the woman with the pearls remained with her family, either because the rumours had not yet reached her, or because she was too proud and fearless to heed them. Tadzio remained. And to Aschenbach, in his infatuation, it seemed at times as though flight and death might remove all the disturbing elements of life around them, and he stay here alone with the boy. Yes, by the sea in the forenoon when his eyes rested heavily, irresponsibly, unwaveringly on the thing he coveted, or when, as the day was ending, he followed shamelessly after him through streets where the hideous death lurked in secret—at such times the atrocious seemed to him rich in possibilities, and laws of morality had dropped away. &0.147 &17.1 &-0.5 &5 &-2.6 \\
\midrule
Human Translator 2 & The stricken man awoke from his dream unnerved and shaken, powerless in the demon’s thrall. He no longer shunned the curious glances of the people around him: whether he aroused their suspicion was of no concern to him. They were fleeing, were they not, leaving Venice: many of the cabanas were vacant, a number of tables in the dining room unoccupied, and in the city there was scarcely a foreigner to be seen. The truth seemed to have leaked out despite the tenacious solidarity among interested parties in their attempt to ward off panic. But the woman in pearls stayed on with her family either because the rumors had not reached her or because she was too proud and fearless to heed them. Thus Tadzio stayed on, and Aschenbach, in his beleaguered state, occasionally had the feeling that flight and death might eliminate all life standing in his way and leave him alone on this island with the beautiful boy; indeed, when he sat in the morning by the sea, his gaze—heavy, injudicious, and fixed—resting on the object of his desire, or when, as evening fell, he resumed his undignified pursuit through the narrow streets clandestinely haunted by loathsome dying, things monstrous seemed auspicious and the moral code null and void. &0.157 &19.1 &-0.4 &6 &-2.6 \\
\midrule
Qwen 2 &From that dream, the wanderer awoke, unsettled, broken, and without strength, surrendering to the demon. He no longer feared the watchful gazes of people; it didn't bother him if they attributed their suspicions. Indeed, they too had fled or left; many beach huts stood empty, the dining hall indicated greater vacuity, and in the city, one seldom saw strangers anymore. The truth seemed to spread, but despite the stubborn unity of the interested parties, panic could not be held back. But the woman adorned with pearls remained, whether shielded from the rumors or too proud and fearless to evade them - Tadzio stayed. In his expanse, it sometimes felt as though flight and death could remove all distractions from the circle, leaving him alone with the Beauty on this island. Indeed, when, in the early morning, his gaze, deliberate, unaccountable, and unwavering, rested on the desired object across the sea, or as he followed the illicit death in hidden streets during descending days, feeling unworthy, he found the abyssal allure irresistible and the moral law. &0.194 &18.6 &-3.4 &3 &-7.6 \\
\midrule
Llama 3 &From this dream, the Heimgesuchte woke up, worn out, shattered, and completely under the demon's control. He no longer feared the observing gazes of people; whether he risked being suspected by them or not, he didn't care. After all, they were fleeing, traveling away; many beach huts stood empty, the occupancy of the dining hall showed larger gaps, and in the city, it was rare to see a stranger. The truth had seeped in, and despite the persistent unity of the interested parties, panic was no longer possible to conceal. But the woman in the pearl jewelry remained with hers, whether the rumors didn't reach her or because she was too proud and fearless to leave them; Tadzio stayed; and to him, in his limited scope, it was sometimes as if flight and death could remove all disturbing life in the circle and leave him alone with the beautiful on this island. - yes, when, in the morning, his gaze, heavy, unaccountable, unattached, rested on the coveted object, when, with the sinking day, he followed unworthily through alleys where concealed, secretly, death was spreading, he seemed to find it promising and futile the law of morals. &0.208 &24.2 &-5 &4 &-7.6 \\
\midrule
TowerInstruct &From this dream the tormented one awoke weary, destroyed, and spiritually weakened to the demon's hold. He no longer shied away from people's gazes; whether he exposed himself to their suspicion did not concern him. Even they, in fact, fled, traveled away; numerous beach huts stood empty, the occupancy of the dining hall showed larger gaps, and one seldom still saw a stranger in the city. The truth seemed to seep through, and the panic, despite the interested party's staunch collective effort, was no longer to be kept at bay. But the woman in pearls stayed with her own, whether the rumors did not reach her, or whether she was too proud and fearless to budge: Tadzio remained; and to that extent, it was as though fleeing and death could all disturbing life around him and allow him alone with the beautiful person on this island to remain—yes, when in the morning by the sea his gaze was heavy, irresponsible, uncharming on the desired one, when he, discreetly, followed him through alleys in which the uncomfortable death circled around him, then it seemed like the monstrous was promising and the moral code fleeting. &0.252 &23.2 &-2.2 &1 &-7.4 \\
\bottomrule
\end{tabular}
  \caption{\label{tab:syntax_example1}
   (Part 1) Example of system outputs showing syntactic similarity to the source and lexical overlap with all other systems. Systems are sorted from low to high syntactic similarity. 
  }
\end{table*}
\begin{table*}
    \centering
    \tiny
\begin{tabular}{C{1cm}m{7cm}C{1cm}C{1cm}C{1cm}C{1cm}C{1cm}}\toprule
\textbf{Model} &\textbf{Translation} &\textbf{Syntactic similarity (Src-Tgt)} &\textbf{Lexical overlap} &\textbf{Human MQM} &\textbf{SQM} &\textbf{\gemba (Literary)} \\
\midrule
Gemma &The troubled awakened from his dream, unnerved, discomposed and adrift before the demon. He no longer feared the watchful eyes of the people; whether he should evade their suspicions mattered not. They were fleeing too, traveling away; many beach shacks stood empty, the dining hall was sparsely populated, and the city held scarcely a stranger. Truth seemed to have drained away, panic struggling to contain itself despite the desperate clinging of the interested parties. But the woman in the pearl jewelry stayed with him, whether the rumors had not reached her, or because she was too proud and fearless to yield to them: Tadzio remained; and in his isolation, it sometimes seemed that flight and death could erase all that was disturbing in life’s round and leave him alone with the beautiful on this island—yes, even if this morning, at the sea’s edge, his gaze hung heavily, irresponsible and untethered, on the beloved one, while behind him, in the whispering gutters, death lingered in silent mockery, making him unworthy of its pursuit. &0.255 &20.2 &-2.6 &4 &-7.6 \\
\midrule
M2M &From this dream the home-seeking man woke up nervous, crushed, and powerlessly fallen to the devil. He no longer feared the observing eyes of the people; whether he exposed himself to their suspicion, he did not care. They also fled, they went away; numerous beach chalets were empty, the occupation of the dining room showed greater gaps, and in the city one rarely saw a stranger. The truth seemed to have leaked, the panic, despite the tight cohesion of the interested parties, no longer holds back. But the woman in the pearl jewelry remained with her own, either because the rumors did not come to her, or because she was too proud and fearless to evade them: Tadzio remained; and to him, in his possession, it was sometimes as if escape and death could remove all the disturbing life in the circle and he remained alone with the Beautiful on this island—yes, when in the morning at the sea his look rested heavy, irresponsible, unrelated to the desired, when he in the sinking day by the streets, in which mysteriously the ugly died, followed him unworthy, so the monstrous seemed to him in sight and fall the law. &0.256 &27.9 &-5 &2 &-7.4 \\
\midrule
\gptfo &Of this dream, the afflicted one awoke unnerved, shattered, and powerless, succumbed to the demon. He no longer shunned the observing gazes of people; whether he exposed himself to their suspicion did not concern him. They too fled, departed; numerous beach huts stood empty, the occupancy of the dining room showed greater gaps, and in the town, one rarely saw a stranger anymore. The truth seemed to have leaked out, and the panic, despite the stubborn cohesion of interests, could no longer be held back. But the woman with the pearl jewelry stayed with her companions, whether because the rumors did not reach her or because she was too proud and fearless to yield to them: Tadzio stayed; and to the one enveloped by his feelings, it sometimes seemed as if flight and death could remove all the disturbing life around them, and he alone would remain with the beautiful one on this island,—yes, when in the mornings by the sea his gaze rested heavily, irresponsibly, unwaveringly on the desired one, when he at the end of the day followed him unworthily through streets where hidden death crept, the monstrous seemed promising, and the moral law obsolete. &0.256 &29.7 &-0.6 &5 &-2.6 \\
\midrule
\gtran &The afflicted man awoke from this dream, unnerved, shattered and powerless, succumbing to the demon. He no longer feared the watching eyes of people; he did not care whether he exposed himself to their suspicions. And they fled, departed; many beach huts stood empty, the dining room was empty and strangers were rarely seen in the town. The truth seemed to have leaked out, and the panic, despite the stubborn solidarity of those interested, could no longer be contained. But the woman in the pearl jewelry stayed with her family, either because the rumors did not reach her or because she was too proud and fearless to give in to them: Tadzio stayed; and to him, in his embarrassment, it sometimes seemed as if flight and death could remove all disturbing life in the area and he alone with the beautiful on this island - yes, when in the morning by the sea his gaze rested heavily, irresponsibly, unflinchingly on the desired one, when as the day was setting he unworthily followed him through streets in which the disgusting death was secretly going around, then the monstrous seemed to him promising and the moral law invalid. &0.257 &30.6 &-0.8 &3 &-7.4 \\
\midrule
NLLB &From this dream the house-boy awoke, nervous, broken, and powerless to fall to the demon. He no longer feared the watching eyes of men; whether he was subject to their suspicion, he did not care. They also fled, they traveled away; numerous beach huts were empty, the occupation of the dining-room showed greater gaps, and in the city one seldom saw a stranger again. The truth seemed to leak through, the panic, despite the tough unity of the interested parties, no longer held back. But the woman in pearl jewelry looked with your eyes, whether it was because the rumors did not press her, or because she was too proud and fearless to reproach them: that one, who remained in captivity, and who, with his mind, could escape and rest for days, when death seemed to him more difficult than the pursuit of the immortal guest, and who, by the unfailing grace of the unknown, returned to the unknown, and so untroubled by the unknown, was able to escape from the shame of the unknown, and to rest with the unforgettable life, as he had done in the unforgettable manner of the unknown. &0.260 &20.5 &-4.4 &1 &-12.4 \\
\midrule
\deepl &The haunted man awoke from this dream exasperated, shattered and powerless, having succumbed to the demon. He no longer shied away from the observing eyes of the people; whether he exposed himself to their suspicion was of no concern to him. They also fled, left; numerous beach huts stood empty, there were large gaps in the dining room staff, and it was rare to see a stranger in the town. The truth seemed to have leaked out and the panic could no longer be contained, despite the tenacious efforts of the interested parties. But the woman in the pearl jewelry stayed with her own, either because the rumors did not reach her or because she was too proud and fearless to give way to them: Tadzio remained; and to him, in his entrapment, it sometimes seemed as if flight and death could remove all disturbing life in the circle, and he could remain alone with the beautiful one on this island,-yes, when in the morning by the sea his gaze rested heavily, irresponsibly, unwaveringly on the desired one, when, as the day sank, he followed him unworthily through alleys where secretly the nasty dying was going on, the monstrous seemed to him promising and the moral law obsolete. &0.261 &31.3 &-2.6 &4 &-2.6 \\
\bottomrule
\end{tabular}
  \caption{\label{tab:syntax_example2}
  (Part 2) Example of system outputs showing syntactic similarity to the source and lexical overlap with all other systems. Systems are sorted from low to high syntactic similarity. 
  }
\end{table*} 
\end{document}